\documentclass{article} 
\usepackage[preprint]{colm2026_conference}

\usepackage{microtype}
\usepackage{hyperref}
\usepackage{url}
\usepackage{amsmath,amssymb}
\usepackage{graphicx}
\usepackage{booktabs}
\usepackage[table]{xcolor}
\usepackage{listings}
\usepackage{pifont}
\usepackage{subcaption}

\usepackage{lineno}
\usepackage{multirow}
\usepackage{enumitem}
\setlist{leftmargin=*, labelsep=0.5em, topsep=2pt, itemsep=1pt}
\usepackage{tikz}
\usetikzlibrary{positioning, arrows.meta, shapes.geometric}
\usepackage[most]{tcolorbox}
\usepackage{wrapfig}
\usepackage{float}

\definecolor{agentbg}{HTML}{F0F4FF}
\definecolor{agentframe}{HTML}{4A6FA5}
\definecolor{toolbg}{HTML}{FFF8E7}
\definecolor{toolframe}{HTML}{C4960C}
\definecolor{errorbg}{HTML}{FFF0F0}
\definecolor{errorframe}{HTML}{CC3333}
\definecolor{successbg}{HTML}{F0FFF0}
\definecolor{successframe}{HTML}{2D8A2D}

\newtcolorbox{agentbox}[1][]{%
    colback=agentbg, colframe=agentframe, fonttitle=\bfseries\small,
    boxrule=0.5pt, arc=2pt, left=4pt, right=4pt, top=2pt, bottom=2pt,
    title={#1}}

\newtcolorbox{toolbox}[1][]{%
    colback=toolbg, colframe=toolframe, fonttitle=\bfseries\small,
    boxrule=0.5pt, arc=2pt, left=4pt, right=4pt, top=2pt, bottom=2pt,
    title={#1}}

\newtcolorbox{errorbox}[1][]{%
    colback=errorbg, colframe=errorframe, fonttitle=\bfseries\small,
    boxrule=0.5pt, arc=2pt, left=4pt, right=4pt, top=2pt, bottom=2pt,
    title={#1}}

\newtcolorbox{successbox}[1][]{%
    colback=successbg, colframe=successframe, fonttitle=\bfseries\small,
    boxrule=0.5pt, arc=2pt, left=4pt, right=4pt, top=2pt, bottom=2pt,
    title={#1}}

\usepackage{tikz}
\usepackage{xcolor}
\usetikzlibrary{positioning,arrows.meta,calc,backgrounds}

\definecolor{tealfill}    {RGB}{225,245,238}
\definecolor{tealstroke}  {RGB}{ 15,110, 86}
\definecolor{tealtext}    {RGB}{  8, 80, 65}
\definecolor{amberfill}   {RGB}{250,238,218}
\definecolor{amberstroke} {RGB}{133, 79, 11}
\definecolor{ambertext}   {RGB}{ 99, 56,  6}
\definecolor{purplefill}  {RGB}{238,237,254}
\definecolor{purplestroke}{RGB}{ 83, 74,183}
\definecolor{purpletext}  {RGB}{ 60, 52,137}
\definecolor{coralfill}   {RGB}{250,236,231}
\definecolor{coralstroke} {RGB}{153, 60, 29}
\definecolor{coraltext}   {RGB}{113, 43, 19}
\definecolor{bluefill}    {RGB}{230,241,251}
\definecolor{bluestroke}  {RGB}{ 24, 95,165}
\definecolor{bluetext}    {RGB}{ 12, 68,124}
\definecolor{greenfill}   {RGB}{234,243,222}
\definecolor{greenstroke} {RGB}{ 59,109, 17}
\definecolor{greentext}   {RGB}{ 39, 80, 10}
\definecolor{grayfill}    {RGB}{241,239,232}
\definecolor{graystroke}  {RGB}{ 95, 94, 90}
\definecolor{graytext}    {RGB}{ 68, 68, 65}
\definecolor{cardbg}      {RGB}{250,250,248}

\def\SC{1.88}    
\def\CW{3.3}     
\def\CTW{3.1}    
\def\yA{-0.90}   
\def\yB{-2.38}   
\def\dropB{1.52} 

\tikzset{
  pstep/.style={rectangle, rounded corners=3pt,
    minimum width=1.6cm, minimum height=0.7cm, align=center,
    font=\sffamily\bfseries\fontsize{6.5}{8}\selectfont,
    inner sep=3pt, line width=0.45pt},
  pstepteal/.style   ={pstep, draw=tealstroke,   fill=tealfill,   text=tealtext},
  pstepamber/.style  ={pstep, draw=amberstroke,  fill=amberfill,  text=ambertext},
  psteppurple/.style ={pstep, draw=purplestroke, fill=purplefill, text=purpletext},
  pstepcoral/.style  ={pstep, draw=coralstroke,  fill=coralfill,  text=coraltext},
  detcard/.style={rectangle, rounded corners=3pt,
    minimum width=\CW cm, text width=\CTW cm,
    align=left, font=\sffamily\fontsize{5.5}{7}\selectfont,
    inner sep=4pt, line width=0.3pt,
    draw=graystroke!45, fill=cardbg, text=graytext},
  xpill/.style={rectangle, rounded corners=4pt,
    font=\sffamily\fontsize{5}{6}\selectfont,
    inner sep=1.8pt, minimum height=0.26cm, line width=0.25pt},
  xpillteal/.style   ={xpill,draw=tealstroke,   fill=tealfill,   text=tealtext},
  xpillamber/.style  ={xpill,draw=amberstroke,  fill=amberfill,  text=ambertext},
  xpillpurple/.style ={xpill,draw=purplestroke, fill=purplefill, text=purpletext},
  xpillcoral/.style  ={xpill,draw=coralstroke,  fill=coralfill,  text=coraltext},
  xpillblue/.style   ={xpill,draw=bluestroke,   fill=bluefill,   text=bluetext},
  xpillgreen/.style  ={xpill,draw=greenstroke,  fill=greenfill,  text=greentext},
  xpillgray/.style   ={xpill,draw=graystroke,   fill=grayfill,   text=graytext},
  mya/.style ={-{Stealth[length=2.8pt,width=2.2pt]},line width=0.45pt,color=graystroke},
  myda/.style={mya,dashed,color=graystroke!50},
  ldr/.style ={dashed,line width=0.25pt,color=graystroke!38},
  gatestyle/.style={rectangle,rounded corners=3pt,align=center,
    font=\sffamily\fontsize{6}{7.5}\selectfont,
    inner sep=5pt,line width=0.4pt,dashed,
    draw=graystroke!65,fill=cardbg,text=graytext},
  mbox/.style={rectangle,rounded corners=3pt,align=center,
    font=\sffamily\fontsize{6}{7.5}\selectfont,
    inner sep=4pt,minimum height=0.82cm,line width=0.4pt},
  mbgreen/.style={mbox,draw=greenstroke,fill=greenfill,text=greentext},
  mbblue/.style ={mbox,draw=bluestroke, fill=bluefill, text=bluetext},
  mbteal/.style ={mbox,draw=tealstroke, fill=tealfill, text=tealtext},
  mbcoral/.style={mbox,draw=coralstroke,fill=coralfill,text=coraltext},
}

\newcommand{\ct}[1]{{\sffamily\bfseries\fontsize{6}{7.5}\selectfont\color{graytext}#1}}
\newcommand{\cb}[1]{{\sffamily\fontsize{5.3}{6.8}\selectfont\color{graytext!80}#1}}
\newcommand{\kw}[2]{{\sffamily\bfseries\fontsize{5.3}{6.8}\selectfont\color{#1text}#2}}

\newcommand{\cmark}{\ding{51}}
\newcommand{\xmark}{\ding{55}}
\newcommand{\aar}{\textsc{AAR}}

\definecolor{darkblue}{rgb}{0, 0, 0.5}
\hypersetup{colorlinks=true, citecolor=darkblue, linkcolor=darkblue, urlcolor=darkblue}

\title{The Amazing Agent Race: Strong Tool Users, Weak Navigators}

\author{Zae Myung Kim$^{1}$, Dongseok Lee$^{2}$, Jaehyung Kim$^{2}$, Vipul Raheja$^{3}$, Dongyeop Kang$^{1}$ \\
        University of Minnesota Twin Cities$^{1}$, Yonsei University$^{2}$, Grammarly$^{3}$\\ 
        \texttt{\{kim01756,dongyeop\}@umn.edu}
}

\begin{document}

\ifcolmsubmission
\linenumbers
\fi

\maketitle

\begin{abstract}
Existing tool-use benchmarks for LLM agents are overwhelmingly \emph{linear}: our analysis of six benchmarks shows 55 to 100\% of instances are simple chains of 2 to 5 steps. We introduce \textsc{The Amazing Agent Race} (\aar{}), a benchmark featuring \emph{directed acyclic graph} (DAG) puzzles (or ``legs'') with fork-merge tool chains. We release 1{,}400 instances across two variants: sequential (800 legs) and compositional (600 DAG legs). Agents must navigate Wikipedia, execute multi-step tool chains, and aggregate results into a verifiable answer. Legs are procedurally generated from Wikipedia seeds across four difficulty levels with live-API validation. Three complementary metrics (finish-line accuracy, pit-stop visit rate, and roadblock completion rate) separately diagnose navigation, tool-use, and arithmetic failures. Evaluating three agent frameworks on 1{,}400 legs, the best achieves only 37.2\% accuracy. Navigation errors dominate (27 to 52\% of trials) while tool-use errors remain below 17\%, and agent architecture matters as much as model scale (Claude Code matches Codex CLI at 37\% with 6$\times$ fewer tokens). The compositional structure of \aar{} reveals that agents fail not at calling tools but at navigating to the right pages, a blind spot invisible to linear benchmarks. The project page can be accessed at: \url{https://minnesotanlp.github.io/the-amazing-agent-race}
\end{abstract}

\section{Introduction}

Consider an innocuous question: ``What is the elevation difference between the
birthplaces of Apple's founders?'' Using Wikipedia as one possible information source, an agent might (1)~navigate to Apple's page,
(2)~extract the founders' names, (3)~follow links to their biographical pages,
(4)~identify their birthplaces (San Francisco and Green Bay), (5)~geocode each city,
(6)~query an elevation API, and (7)~compute the difference:

\vspace{-0.5em}
{\small
\begin{verbatim}
  coords_1 = geocode("San Francisco") → (37.77, -122.42)
  coords_2 = geocode("Green Bay")     → (44.51, -88.01)
  elev_1   = elevation(coords_1)      → 16 m
  elev_2   = elevation(coords_2)      → 177 m
  answer   = abs(elev_1 - elev_2)     → 161 m
\end{verbatim}
}
\vspace{-0.5em}

\noindent A wrong page visit or swapped coordinate cascades through the chain and invalidates the answer. If the question also asks for the driving distance, the agent must \emph{fork} coordinates into parallel API calls and \emph{merge} results, a non-linear dependency that existing benchmarks leave untested.

Existing benchmarks isolate these capabilities: tool-use benchmarks~\citep{qin2024toolllm,patil2025bfcl} omit navigation, compositional benchmarks~\citep{basu2024nestful,ye2025toolhop} provide all inputs upfront, and web-navigation benchmarks~\citep{zhou2024webarena,mialon2023gaia} omit compositional tool chains. Our analysis of their dependency structures reveals that 55 to 100\% of instances are strictly linear chains averaging only 2 to 5 steps (\S\ref{sec:related}), a \emph{compositionality deficit} that leaves fork--merge reasoning untested.

\noindent\textbf{This work.}
We introduce \textsc{The Amazing Agent Race} (\aar{}), a benchmark designed around
one diagnostic question: \emph{where exactly does an agent break down when it must
discover information through navigation, fork that information into parallel tool
branches, and merge the results?} Inspired by the television series \emph{The
Amazing Race}~\citep{amazingrace}, \aar{} frames evaluation as a race across
Wikipedia. Each instance is a \emph{leg}: a sequence of steps where the agent navigates
Wikipedia pages, executes tool chains (e.g., geocode $\to$ elevation, geocode $\to$
weather), applies analytical reasoning, and aggregates results into a single-digit
answer. Legs are not linear chains but directed acyclic graphs (DAGs): fork--merge
\emph{diamond} patterns spawn parallel tool branches from a single extracted entity
whose outputs merge downstream. Every \aar{} instance is a true DAG (0\% linear)
with an average of 22 pit stops and up to 5 diamonds, compared to 94--100\%
linearity and 1.7--4.8 steps in prior benchmarks.

An automated pipeline generates legs from random Wikipedia seeds with pre-validated tool chains, diamond augmentation, and verbalized clue envelopes that never reveal titles or tool names directly. \aar{} provides 19 tools across four difficulty levels (8 to 33 pit stops); live APIs ensure answers must be \emph{derived}, not recalled.

\begin{figure}[t]
\centering
\includegraphics[width=\textwidth]{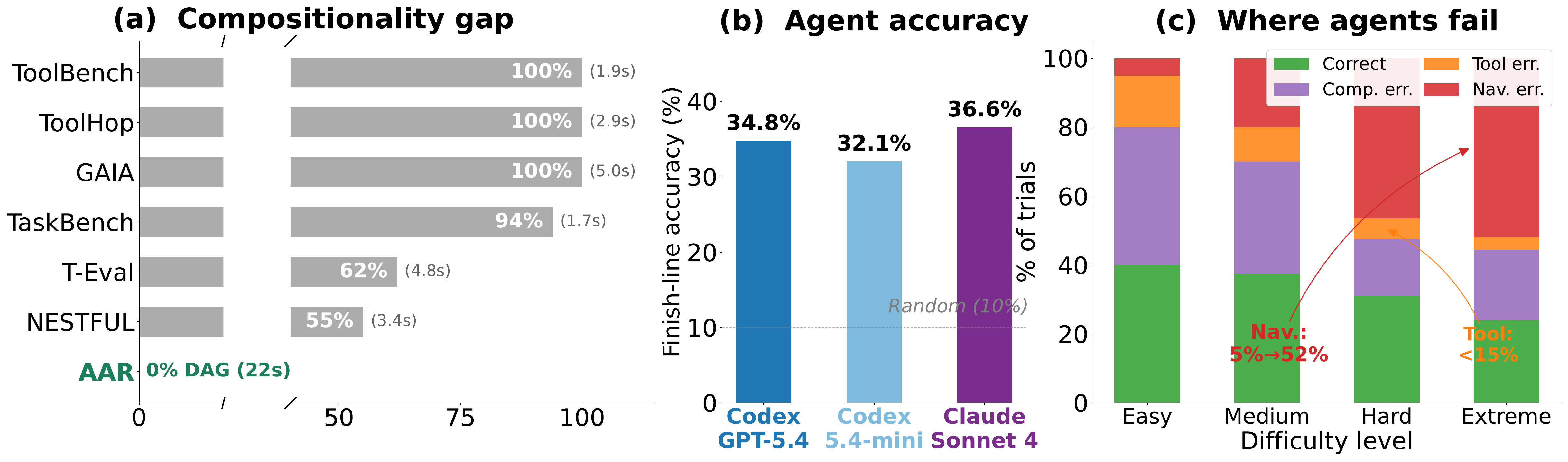}
\caption{
  \textbf{(a)} Existing benchmarks are 55 to 100\% linear; \aar{} is 0\% linear (all DAGs). Numbers in parentheses show mean steps per instance (abbreviated ``s'').
  \textbf{(b)} Best agent accuracy is 36.6\% (aggregated across 1{,}400 legs).
  \textbf{(c)} Navigation errors dominate (5\% to 52\%) while tool-use errors stay below 15\%.
}
\vspace{-6mm}
\label{fig:motivation}
\end{figure}

Three metrics separately diagnose failures at each pipeline stage (Figure~\ref{fig:motivation}): finish-line accuracy (FA), pit-stop visit rate (PVR, navigation), and roadblock completion rate (RCR, tool use).

\noindent\textbf{Key findings.}
Evaluating three agent frameworks on 1{,}400 legs, the best achieves only 37.2\% FA. Navigation errors dominate (27 to 52\% of trials) while tool-use errors stay below 17\%. Moving from \aar{}-Linear to \aar{}-DAG drops navigation scores by 13 to 18pp while tool-use scores remain stable, confirming that compositional structure challenges navigation, not tool use (\S\ref{sec:main_results}).

\noindent\textbf{Contributions.}
\begin{enumerate}[nosep]
  \item A \emph{compositionality analysis} of six benchmarks showing 55--100\%
  linearity (\S\ref{sec:related}).
  \item An \emph{automated generation pipeline} producing DAG-structured legs from
  random Wikipedia seeds with fork--merge diamond patterns, four structurally
  controlled difficulty levels, and contamination resistance via live APIs and clue
  paraphrasing (\S\ref{sec:pipeline}--\S\ref{sec:diamonds}).
  Code and data are available at \url{https://github.com/minnesotanlp/the-amazing-agent-race}.
  \item \emph{Three decomposed metrics} (FA, PVR, RCR) that isolate failures at the
  navigation, tool-use, and computation stages (\S\ref{sec:experiments}). \emph{Evaluation on 1,400 legs} across three agent frameworks and two model
  families, with a detailed failure taxonomy (\S\ref{sec:main_results},
  \S\ref{sec:discussion}).
\end{enumerate}

\section{Related Work}
\label{sec:related}

\begin{table}[t]
\centering
\scriptsize
\setlength{\tabcolsep}{2.5pt}
\renewcommand{\arraystretch}{1.0}
\begin{tabular}{@{}lcccccccccccc@{}}
\toprule
& & & \multicolumn{4}{c}{\textit{Evaluation}} &
  \multicolumn{3}{c}{\textit{Design}} &
  \multicolumn{3}{c}{\textit{Compositionality}} \\
\cmidrule(lr){4-7}\cmidrule(lr){8-10}\cmidrule(lr){11-13}
\textbf{Benchmark} & \textbf{Venue} & \textbf{Tools}
  & \textbf{Nav} & \textbf{Met} & \textbf{Stp} & \textbf{Lve}
  & \textbf{Diff} & \textbf{Gld} & \textbf{Gen}
  & \textbf{Steps} & \textbf{\%Lin} & \textbf{\%DAG} \\
\midrule
\multicolumn{13}{l}{\textit{Tool-use \& composition}} \\
ToolBench    & ICLR'24   & 16k+  & \xmark & 2 & \xmark & \cmark$^\dagger$
             & 3 lvl & \cmark & Auto  & 1.9  & 100 & 0   \\
TaskBench    & NeurIPS'24 & graph & \xmark & 3 & \cmark & \xmark
             & size  & \cmark & Auto  & 1.7  & 94  & 2.5 \\
NESTFUL      & arXiv'24  & nest  & \xmark & 2 & \cmark & \xmark
             & depth & \cmark & Scr   & 3.4  & 55  & 45  \\
\midrule
\multicolumn{13}{l}{\textit{Web navigation \& agent}} \\
GAIA         & ICLR'24   & var   & \cmark & 1 & \xmark & \xmark
             & 3 lvl & \xmark & Man   & $\sim$5$^\ddagger$ & 100 & 0 \\
WebArena     & ICLR'24   & brow  & \cmark & 1 & \xmark & \cmark
             & impl  & \xmark & Scr   & --   & --  & --  \\
AgentBench   & ICLR'24   & 8env  & part  & 1 & \cmark & mix
             & env   & \xmark & Man   & --   & --  & --  \\
\midrule
\textbf{\aar{}} & --     & \textbf{19} & \cmark & \textbf{3} & \cmark & \cmark
             & \textbf{4 lvl} & \cmark & \textbf{Auto} & \textbf{22.1} & \textbf{0} & \textbf{100} \\
\bottomrule
\end{tabular}
\caption{Comparison with representative benchmarks (3 per category; full table with 12 benchmarks in Appendix~\ref{appendix:full_comparison}).
$^\dagger$ToolBench suffers API instability.
$^\ddagger$GAIA step count from annotator metadata only.}
\label{tab:comparison}
\vspace{-6mm}
\end{table}

Deploying an LLM agent in the wild requires interpreting instructions, navigating
information sources, invoking APIs, and chaining results, all within a single
episode. Existing benchmarks isolate one or two of these capabilities; \aar{}
combines open web navigation with multi-step tool composition in a structurally
controlled, automatically generated benchmark (Table~\ref{tab:comparison}).

\noindent\textbf{Tool-use benchmarks.}
ToolBench~\citep{qin2024toolllm} curates 16{,}464 REST APIs for multi-step
planning; real-API instability motivated StableToolBench~\citep{guo2024stabletoolbench}
to replace live endpoints with a virtual server. BFCL~\citep{patil2025bfcl}
standardizes function-calling evaluation with AST-based scoring and multi-turn
stateful workflows. API-Bank~\citep{li2023apibank} introduces a three-level
framework over 73 APIs. All three scale the \emph{number} of available tools but
present them in isolation: the agent receives a query and calls APIs without needing
to \emph{find} the inputs first.

\noindent\textbf{Multi-step tool composition.}
TaskBench~\citep{shen2024taskbench} models inter-tool dependencies as a Tool Graph.
NESTFUL~\citep{basu2024nestful} tests nested API sequences (GPT-4o: 28\%
full-sequence accuracy). ToolHop~\citep{ye2025toolhop} constructs multi-hop queries
requiring 3+ chained calls (best model: 49\%). T-Eval~\citep{chen2024teval}
decomposes tool use into six sub-capabilities. ToolSandbox~\citep{lu2025toolsandbox}
adds statefulness and implicit dependencies. These benchmarks show compositional
tool use is hard even when all inputs are given upfront. \aar{} adds a further
challenge: agents must first \emph{discover} inputs through navigation, coupling
navigation errors with downstream tool failures.

\noindent\textbf{Compositionality gap.}
We extract dependency graphs from the golden execution traces of six benchmarks
(Table~\ref{tab:comparison}). ToolBench, ToolHop, and GAIA are entirely linear
(100\%). TaskBench, the only benchmark with explicit DAG annotations, is 94\%
linear with just 1.7 steps on average. NESTFUL and T-Eval show moderate
non-linearity (45\% and 38\%) but remain shallow (3.4 and 4.8 steps). Every \aar{}
instance is a DAG averaging 22 pit stops with fan-out and fan-in through diamond
patterns, a structural gap that motivates our benchmark.\footnote{GAIA lacks
structured golden chains; we use annotator-reported step counts as a linear-chain
proxy (165 validation samples only).}

\noindent\textbf{Web navigation benchmarks.}
WebArena~\citep{zhou2024webarena} evaluates long-horizon tasks across self-hosted
web applications. Mind2Web~\citep{deng2024mind2web} tests generalization across 137
real websites. OSWorld~\citep{xie2024osworld} extends evaluation to desktop GUI
environments. GAIA~\citep{mialon2023gaia} comes closest to \aar{}'s scope (some
questions require both web lookup and tool use), but its 466 manually curated,
static instances risk contamination, difficulty is human-annotated rather than
structurally controlled, and evaluation is limited to final-answer exact match.
\aar{} addresses all three limitations.

\noindent\textbf{Broader context.}
Holistic multi-environment benchmarks~\citep{liu2023agentbench,ma2024agentboard,trivedi2024appworld,yao2024taubench,xu2024theagentcompany} trade depth for breadth; \aar{} makes the complementary trade-off. Contamination resistance via live APIs and procedural generation is discussed alongside related fixed-benchmark limitations in Appendix~\ref{appendix:more_related_work}.

\vspace{-5mm}

\section{Benchmark Design Principles}
\label{sec:design}

While our framework is source-agnostic, we use Wikipedia because it offers dense hyperlink graphs ($\sim$40 outgoing links per page), semi-structured infoboxes for deterministic fact extraction, broad topical diversity, free licensing (CC BY-SA), and a contamination testbed: since LLMs have trained extensively on Wikipedia, our benchmark specifically tests whether agents can go \emph{beyond} memorized facts via paraphrased clues and live API calls (\S\ref{sec:quality}).

\begin{figure}[t]
\centering
\includegraphics[width=\columnwidth]{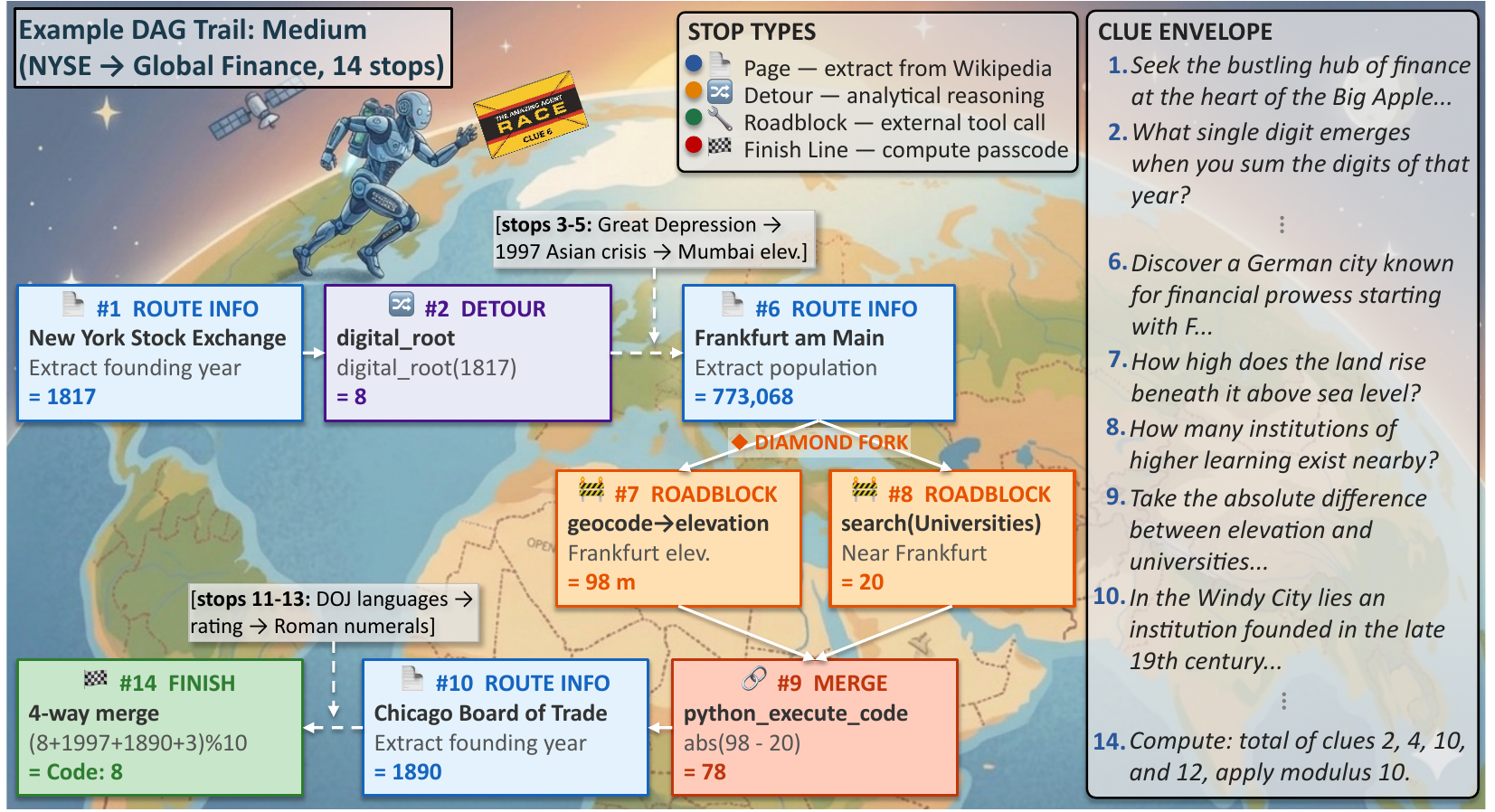}
\caption{An example clue envelope (or a ``leg'') as presented to the agent.}
\label{fig:riddle_example}
\vspace{-5mm}
\end{figure}

\vspace{-3mm}
\subsection{Task Formulation}
\label{sec:task}
\vspace{-3mm}

An \aar{} instance (a \emph{leg}) consists of four inputs and produces one output:
 
\begin{itemize}[nosep]
  \item A \emph{seed URL} $u_0$ pointing to a Wikipedia article (the starting line).
  \item A \emph{clue envelope} $\mathcal{C}$: a natural-language riddle whose $K$ clues describe a sequence of steps without naming Wikipedia titles or tool names.
  \item A \emph{tool set} $\mathcal{T}$ of 19 tools with schema descriptions.
  \item A \emph{step budget} $B = \max(10,\lfloor 1.5K \rfloor)$.
\end{itemize}
 
\noindent The agent must produce a single-digit \emph{finish-line code} $\hat{y}\in\{0,\dots,9\}$. The ground-truth code $y^{*}$ is computed by the golden executor from a verified execution trace.

\vspace{-3mm}
\subsection{Leg Structure}
\label{sec:leg_structure}
\vspace{-2mm}

A leg is a directed acyclic graph (DAG) of \textbf{pit stops} $s_1,\dots,s_K$, each producing a typed value $v_i$ and optionally depending on prior stops via explicit \texttt{depends\_on} edges. Borrowing terminology from \emph{The Amazing Race}~\citep{amazingrace}, we define four pit-stop types:
 
\begin{enumerate}[nosep]
  \item \textbf{Route info} (\texttt{route\_info}): Navigate to a Wikipedia page and extract a fact (e.g., a numeric infobox field, a date from prose).
  \item \textbf{Roadblock} (\texttt{roadblock}): Execute a multi-step tool chain, e.g., geocode a location then query the elevation API.
  \item \textbf{Detour} (\texttt{detour}): Apply an analytical transform to a prior value, e.g., $\texttt{next\_prime}(v_i)$, $\texttt{digit\_sum}(v_i)$.
  \item \textbf{Finish line} (\texttt{finish\_line}): Aggregate values from earlier stops via arithmetic to produce $y^{*}\in\{0,\dots,9\}$.
\end{enumerate}
 
\noindent Transitions are typed (\texttt{link\_follow}, \texttt{search\_query}, \texttt{tool\_call}, \texttt{compute}), and values are typed (\texttt{number}, \texttt{text}, \texttt{coords}, \texttt{date}), enabling type-aware argument passing between stops.

\vspace{-3mm}
\subsection{Diamond Patterns}
\label{sec:diamonds}
\vspace{-2mm}

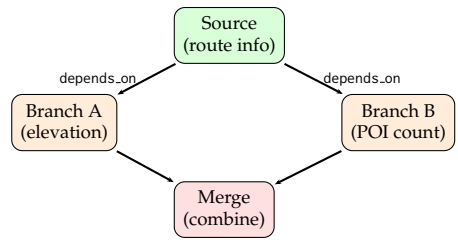
\begin{wrapfigure}[9]{r}{0.44\textwidth}
\centering\vspace{-8mm}
\begin{tikzpicture}[
    node distance=0.4cm and 0.75cm,
    every node/.style={font=\small},
    stop/.style={rectangle, draw, rounded corners, minimum width=1.1cm, minimum height=0.35cm, align=center, font=\scriptsize},
    arr/.style={-{Stealth[length=2.5pt]}, thick}
]
\node[stop, fill=green!15] (src) {Source\\(route info)};
\node[stop, fill=orange!15, below left=of src] (ba) {Branch A\\(elevation)};
\node[stop, fill=orange!15, below right=of src] (bb) {Branch B\\(POI count)};
\node[stop, fill=red!12, below right=of ba] (merge) {Merge\\(combine)};

\draw[arr] (src) -- node[left, font=\tiny] {\texttt{depends\_on}} (ba);
\draw[arr] (src) -- node[right, font=\tiny] {\texttt{depends\_on}} (bb);
\draw[arr] (ba) -- (merge);
\draw[arr] (bb) -- (merge);
\end{tikzpicture}
\caption{Diamond pattern structure.}
\label{fig:diamond}
\end{wrapfigure}

\aar{} introduces \textbf{diamond patterns} (Figure~\ref{fig:diamond}) to create non-linear DAG structure. A diamond has a \emph{source stop} (extract a geocodable entity), two \emph{branch stops} (independent tool chains on the same entity, e.g., elevation and POI count), and a \emph{merge stop} (combines branch outputs). Each branch records a \texttt{depends\_on} edge to the source; the merge depends on both branches.
Diamond count scales with difficulty (1 for easy up to 3--5 for extreme) across four types (\texttt{elevation$\times$POI}, \texttt{elevation$\times$rating}, \texttt{population$\times$area}, \texttt{temperature$\times$precipitation}), guaranteeing every instance is a true DAG.

\subsection{Tool Set}
\label{sec:tools}

\aar{} provides 19 tools across eight categories (Appendix~\ref{appendix:templates}), designed for composability (e.g., \texttt{geocode} $\to$ \texttt{elevation}) and temporal dynamism (stock/crypto tools return live data). Roadblock pit stops instantiate 17 templates composing 1--3 tools. Each tool returns values in a canonical unit (elevation in meters, distance in km, temperature in \textdegree{}C); explicit \texttt{python\_execute\_code} conversion stops handle unit changes when needed. The finish-line stop reduces gathered values to a single digit via modular arithmetic ($\texttt{digital\_root}$, $\texttt{mod10}$, etc.), absorbing small API perturbations.

\subsection{Difficulty Levels}
\label{sec:difficulty}

Difficulty is controlled through four levels that independently vary five parameters: pre-augmentation leg length (3--6 for easy up to 17--21 for extreme), roadblock count, detour count, extraction complexity (infobox-only vs.\ cross-section), and Wikipedia crawl depth (1--3 hops). After diamond augmentation (\S\ref{sec:diamonds}), each diamond adds 3 stops, so final pit-stop counts exceed the configured ranges (e.g., extreme legs average 33 stops from a configured range of 17--21). Higher difficulty simultaneously increases interaction depth along multiple axes. Full parameter ranges are in Table~\ref{tab:difficulty} (Appendix~\ref{appendix:difficulty}).

\section{The \aar{} Benchmark Construction}\label{sec:pipeline}

\def\SC{1.88}
\def\yA{-0.72}
\def\yB{-0.72}
\def\dropB{0.55}
\renewcommand{\ct}[1]{{\sffamily\bfseries\fontsize{7}{8}\selectfont #1}}
\renewcommand{\cb}[1]{{\sffamily\fontsize{5.8}{6.8}\selectfont #1}}
\renewcommand{\kw}[2]{{\sffamily\bfseries\fontsize{5.8}{6.8}\selectfont\color{#1text}#2}}
\tikzset{
pstepteal/.style={rectangle,rounded corners=2pt,draw=teal!70!black,fill=teal!18,
font=\sffamily\bfseries\fontsize{8}{9}\selectfont,
inner sep=3pt,outer sep=0pt,minimum width=1.55cm,minimum height=0.52cm},
pstepamber/.style={rectangle,rounded corners=2pt,draw=amberstroke,fill=amberfill,
font=\sffamily\bfseries\fontsize{8}{9}\selectfont,
inner sep=3pt,outer sep=0pt,minimum width=1.55cm,minimum height=0.52cm},
psteppurple/.style={rectangle,rounded corners=2pt,draw=violet!70!black,fill=violet!18,
font=\sffamily\bfseries\fontsize{8}{9}\selectfont,
inner sep=3pt,outer sep=0pt,minimum width=1.55cm,minimum height=0.52cm},
pstepcoral/.style={rectangle,rounded corners=2pt,draw=coralstroke,fill=coralfill,
font=\sffamily\bfseries\fontsize{8}{9}\selectfont,
inner sep=3pt,outer sep=0pt,minimum width=1.55cm,minimum height=0.52cm},
detcard/.style={rectangle,draw=black!50,fill=white,rounded corners=2pt,
text width=1.45cm,inner sep=2.5pt,outer sep=0pt,align=left,
font=\sffamily\fontsize{5.8}{6.8}\selectfont},
gatestyle/.style={rectangle,draw=black!65,fill=black!5,dashed,
text width=14.2cm,inner sep=3pt,outer sep=0pt,align=left,
font=\sffamily\fontsize{6.2}{7.2}\selectfont},
mbblue/.style={mbox,draw=bluestroke,fill=bluefill,text=bluetext},
mbteal/.style={mbox,draw=tealstroke,fill=tealfill,text=tealtext},
mbcoral/.style={mbox,draw=coralstroke,fill=coralfill,text=coraltext},
xpillblue/.style={rectangle,rounded corners=1pt,draw=blue!70!black,fill=blue!15,
font=\sffamily\fontsize{4.5}{5}\selectfont,inner sep=1pt,outer sep=0pt},
mya/.style={->,>=stealth,line width=1.0pt,color=black},
myda/.style={->,>=stealth,line width=0.8pt,dashed,color=black},
ldr/.style={-,line width=0.6pt,color=black},
}
\begin{figure}[t]
\centering
\resizebox{\textwidth}{!}{%
\begin{tikzpicture}[node distance=0pt]
\node[pstepteal] (s1) at (0*\SC cm,0) {Crawl};
\node[pstepteal] (s2) at (1*\SC cm,0) {Plan};
\node[pstepteal] (s3) at (2*\SC cm,0) {Build};
\node[pstepamber] (s4) at (3*\SC cm,0) {Validate};
\node[pstepteal] (s5) at (4*\SC cm,0) {Link};
\node[psteppurple] (s6) at (5*\SC cm,0) {Augment};
\node[pstepteal] (s7) at (6*\SC cm,0) {Execute};
\node[pstepcoral] (s8) at (7*\SC cm,0) {Verbalize};
\foreach \a/\b in {s1/s2,s2/s3,s3/s4,s4/s5,s5/s6,s6/s7,s7/s8}{
\draw[mya] (\a.east)--(\b.west);
}
\draw[myda] (s8.north)--++(0,0.28)
--($(s1.north)+(0,0.28)$)--(s1.north);
\node[font=\sffamily\fontsize{5.8}{7}\selectfont\color{black!55},
above=0.32cm of s4,xshift=0.6cm]
{fail $\to$ discard \& regenerate};
\node[detcard,anchor=north] (c1) at (0*\SC cm,\yA cm){%
\ct{WikiGraph}\\[0.5pt]
\cb{Outgoing links\\1--3 hops\\Cache: infobox}};
\draw[ldr] (s1.south)--(s1.south|-c1.north);
\node[detcard,anchor=north] (c2) at (1*\SC cm,\yB cm){%
\ct{LLM planner}\\[0.5pt]
\cb{Theme + stops\\Extraction hints\\Difficulty params}};
\draw[ldr] (s2.south)--(s2.south|-c2.north);
\node[detcard,anchor=north] (c3) at (2*\SC cm,\yA cm){%
\ct{LegBuilder}\\[0.5pt]
\cb{\kw{blue}{route info}\\\kw{coral}{roadblock}\\\kw{purple}{detour}}}
;
\draw[ldr] (s3.south)--(s3.south|-c3.north);
\node[detcard,anchor=north] (c4) at (3*\SC cm,\yB cm){%
\ct{Pre-validate}\\[0.5pt]
\cb{Dry-run chains\\Re-index leg\\Solvability chk}};
\draw[ldr] (s4.south)--(s4.south|-c4.north);
\node[detcard,anchor=north] (c5) at (4*\SC cm,\yA cm){%
\ct{Route linker}\\[0.5pt]
\cb{Connect stops\\\texttt{link\_follow}\\\texttt{search\_query}}};
\draw[ldr] (s5.south)--(s5.south|-c5.north);
\node[detcard,text width=1.6cm,anchor=north] (c6) at (5*\SC cm,\yB cm){%
\ct{Diamond\-Augmenter}\\[0.5pt]
\cb{Fork-merge DAG\\\texttt{depends\_on}\\1--5 diamonds}};
\draw[ldr] (s6.south)--(s6.south|-c6.north);
\node[detcard,text width=1.65cm,anchor=north] (c7) at (6*\SC cm,\yA cm){%
\ct{GoldenExecutor}\\[0.5pt]
\cb{Dep.-order run\\Compute $y^*$\\Cache trace}};
\draw[ldr] (s7.south)--(s7.south|-c7.north);
\node[detcard,text width=1.65cm,anchor=north] (c8) at (7*\SC cm,\yB cm){%
\ct{ClueVerbalizer}\\[0.5pt]
\cb{Title paraphrase\\3-pass correction\\Align $\geq$0.7}};
\draw[ldr] (s8.south)--(s8.south|-c8.north);
\node[gatestyle,anchor=north]
(gate) at (3.5*\SC cm,-2.6cm)
{\textbf{Validation gate}\enskip
Alignment $\geq 0.7$ $\cdot$
Implied code $= y^*$ $\cdot$
No direct Wikipedia titles
$\Rightarrow$ \textit{fail}: discard \& regenerate};
\foreach \c in {c1,c2,c3,c4,c5,c6,c7,c8}{
\draw[ldr,opacity=0.35] (\c.south)--(\c.south|-gate.north);
}

\node[rectangle,draw=teal!60!black,fill=teal!12,rounded corners=2pt,
text width=2.1cm,inner sep=3pt,outer sep=0pt,align=center,
font=\sffamily\fontsize{7}{8.5}\selectfont,anchor=west]
(verified) at (0.8cm,-3.65cm)
{\textbf{Verified leg}\\
{\fontsize{5.5}{7}\selectfont 6 quality invariants}};
\node[mbblue,minimum width=2.8cm,anchor=west]
(fa) at ($(verified.east)+(0.22,0)$)
{\textbf{Finish-line acc.\ (FA)}\\
{\fontsize{5}{6.5}\selectfont $\hat{y}=y^*$}};
\node[mbteal,minimum width=2.9cm,anchor=west]
(pvr) at ($(fa.east)+(0.22,0)$)
{\textbf{Pit-stop visit (PVR)}\\
{\fontsize{5}{6.5}\selectfont Nav.\ quality}};
\node[mbcoral,minimum width=3.1cm,anchor=west]
(rcr) at ($(pvr.east)+(0.22,0)$)
{\textbf{Roadblock compl.\ (RCR)}\\
{\fontsize{5}{6.5}\selectfont Tool-use competence}};
\draw[mya] (verified.north|-gate.south)--(verified.north);
\draw[mya] (verified.east)--(fa.west);
\draw[mya] (fa.east)--(pvr.west);
\draw[mya] (pvr.east)--(rcr.west);
\node[font=\sffamily\bfseries\fontsize{6}{7}\selectfont\color{graytext},
anchor=west] (lbl1) at ($(verified.south west)+(0,-0.2)$) {Step:};
\node[xpillteal, anchor=west,right=2pt of lbl1] (la) {data proc.};
\node[xpillamber, anchor=west,right=2pt of la] (lb) {quality chk};
\node[xpillpurple,anchor=west,right=2pt of lb] (lc) {DAG struct.};
\node[xpillcoral, anchor=west,right=2pt of lc] (ld) {NL gen.};
\node[font=\sffamily\bfseries\fontsize{6}{7}\selectfont\color{graytext},
anchor=west,right=8pt of ld] (lbl2) {Pit-stop:};
\node[xpillteal, anchor=west,right=2pt of lbl2] (le) {route info};
\node[xpillamber, anchor=west,right=2pt of le] (lf) {roadblock};
\node[xpillpurple,anchor=west,right=2pt of lf] (lg) {detour};
\node[xpillcoral, anchor=west,right=2pt of lg] (lh) {finish line};
\end{tikzpicture}%
}
\caption{The eight-step automated pipeline for generating \aar{} benchmark legs.
Each leg passes a validation gate before producing evaluation targets:
finish-line accuracy~(FA), pit-stop visit rate~(PVR), and roadblock completion rate~(RCR).}
\label{fig:pipeline}
\vspace{-5mm}
\end{figure}
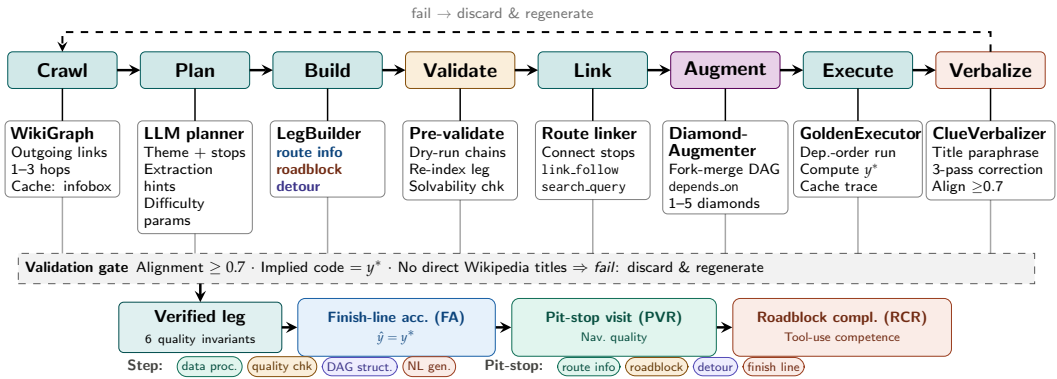

\subsection{Automated Generation Pipeline}

Each leg is produced through an eight-step automated pipeline:

\begin{enumerate}[nosep]
\item \textbf{Crawl.} Fetch the seed page and follow outgoing links, caching infobox fields and content.
\item \textbf{Plan.} Plan a thematic route with pit-stop extraction hints subject to difficulty parameters.
\item \textbf{Build.} Instantiate concrete stops: route-info (fact extraction), roadblocks (tool-chain templates), and detours (analytical transforms).
\item \textbf{Pre-validate.} Dry-run every tool chain against live APIs; drop failing chains and re-index.
\item \textbf{Link.} Connect consecutive stops via \texttt{link\_follow} or \texttt{search\_query}.
\item \textbf{Augment.} Insert the diamond patterns (\S\ref{sec:diamonds}), transforming the chain into a DAG.
\item \textbf{Execute.} Run all chains in dependency order, computing ground-truth values and $y^{*}$.
\item \textbf{Verbalize.} Convert to a clue envelope with circumlocutions (no direct Wikipedia titles). Accept only when round-trip alignment $\geq 0.7$ and implied code $= y^{*}$.
\end{enumerate}

\subsection{Quality Assurance and Contamination Resistance}
\label{sec:quality}

Every leg satisfies six invariants: solvability (golden executor produces $y^*$), API stability (dry-run at generation time), reproducibility (cached traces and page snapshots), input cleanliness, geocodability filtering, and clue-envelope integrity (round-trip alignment $\geq 0.7$, no direct Wikipedia titles).

\aar{} resists memorization through four mechanisms: (1)~clue paraphrasing replaces titles with circumlocutions, (2)~roadblock answers depend on live APIs whose values change, (3)~detour transforms produce values absent from Wikipedia, and (4)~finish-line codes use modular arithmetic over procedurally generated instances. Full details are in Appendix~\ref{appendix:validity}.

\subsection{The \aar{} Dataset: 1{,}400 legs}
\setlength{\columnsep}{12pt}%
\begin{wraptable}[16]{r}{0.49\textwidth}
\centering
\footnotesize\vspace{-10mm}
\setlength{\tabcolsep}{3pt}
\begin{tabular}{@{}ll ccccc@{}}
\toprule
\textbf{Var.} & \textbf{Level} & \textbf{Legs} & \textbf{Stops} & \textbf{RB} & \textbf{Det.} & \textbf{Tools} \\
\midrule
\multirow{5}{*}{\parbox{1.1cm}{\textbf{\aar{}-\\Lin.}}}
& Easy    & 200 & 5.4  & 1.1 & 1.0 & 1.4 \\
& Medium  & 200 & 11.7 & 2.3 & 2.3 & 2.9 \\
& Hard    & 200 & 18.7 & 4.1 & 3.9 & 5.0 \\
& Extreme & 200 & 24.3 & 5.4 & 5.3 & 6.6 \\
& \cellcolor{gray!10}\textit{All} & \cellcolor{gray!10}\textit{800} & \cellcolor{gray!10}\textit{15.0} & \cellcolor{gray!10}\textit{3.2} & \cellcolor{gray!10}\textit{3.1} & \cellcolor{gray!10}\textit{4.0} \\
\midrule
\multirow{5}{*}{\parbox{1.1cm}{\textbf{\aar{}-\\DAG}}}
& Easy    & 100 & 8.3  & 3.6 & 1.0 & 4.8 \\
& Medium  & 150 & 15.4 & 6.0 & 2.3 & 7.9 \\
& Hard    & 166 & 24.6 & 10.1 & 3.9 & 13.2 \\
& Extreme & 184 & 33.0 & 14.1 & 5.2 & 18.2 \\
& \cellcolor{gray!10}\textit{All} & \cellcolor{gray!10}\textit{600} & \cellcolor{gray!10}\textit{22.1} & \cellcolor{gray!10}\textit{9.2} & \cellcolor{gray!10}\textit{3.4} & \cellcolor{gray!10}\textit{12.0} \\
\midrule
\multicolumn{2}{l}{\textbf{Total}} & \textbf{1{,}400} & & & & \\
\bottomrule
\end{tabular}
\caption{Dataset statistics. \textbf{Stops}: mean per leg. \textbf{RB}: roadblocks. \textbf{Det.}: detours. \textbf{Tools}: tool invocations in the golden trace.}
\label{tab:dataset_stats}
\end{wraptable}
We release two benchmark variants (Table~\ref{tab:dataset_stats}): \textbf{\aar{}-Linear} (800 legs with sequential tool chains, 200 per difficulty level) and \textbf{\aar{}-DAG} (600 legs with diamond fork--merge patterns). 
Both are generated from random Wikipedia seed articles sampled from the top 100{,}000 most-viewed English pages. Each leg passes the full quality pipeline: tool-chain pre-validation, golden execution, diamond augmentation (DAG only), and round-trip clue-envelope validation (\S\ref{sec:quality}). Legs that fail any stage are discarded and regenerated.
Every leg is verified solvable by the golden executor, and inter-instance diversity is high (mean pairwise Jaccard similarity of 0.0005 across 10K sampled pairs). Temporal stability is ensured by caching golden traces and using modular arithmetic that absorbs small API perturbations. Full validity analyses are in Appendix~\ref{appendix:validity}.

\vspace{-3mm}
\section{Experimental Setup}
\label{sec:setup}
\vspace{-3mm}

\noindent\textbf{Evaluation framework.}
We run all evaluations through \textbf{Harbor}~\citep{harbor2026}, an open-source agent evaluation framework that orchestrates trials in containerized Docker environments. Harbor wraps diverse agent implementations behind a common interface, enabling fair comparison: each agent receives the same Docker environment with a command-line tool executor (\texttt{tools.py}) that provides access to all 19 \aar{} tools, the clue envelope as a Markdown instruction file, and internet access for web fetching. The agent must write its single-digit answer to \texttt{/app/answer.txt}. A verifier then compares the answer against the golden finish-line code and computes partial-credit metrics by analyzing the agent's tool-call logs against the golden execution trace.

\noindent\textbf{Agent frameworks.}
We evaluate three agent architectures to test whether \aar{} discriminates along architectural lines: \textbf{Codex CLI}: OpenAI's agentic coding assistant with autonomous planning, shell execution, and tool-use capabilities, \textbf{Claude Code}: Anthropic's agentic coding assistant, which autonomously plans, executes shell commands, and iterates on errors, and \textbf{mini-swe-agent}: A lightweight SWE-agent variant supporting multi-step tool orchestration via a ReAct-style bash loop.

\noindent\textbf{Models.}
Codex CLI and mini-swe-agent are evaluated with two OpenAI models (GPT-5.4 and GPT-5.4-mini), Claude Code uses Anthropic's Claude Sonnet~4, and we additionally evaluate Codex CLI with an open-weight reasoning model (GPT-OSS-120B, served via OpenRouter\footnote{\url{https://openrouter.ai/openai/gpt-oss-120b}}):
\textbf{GPT-5.4}: Frontier-scale OpenAI model,
\textbf{GPT-5.4-mini}: Cost-efficient OpenAI variant,
\textbf{Claude Sonnet 4}: Anthropic's frontier model, and
\textbf{GPT-OSS-120B}: Open-weight reasoning model with extended thinking, testing whether reasoning-optimized models can compensate for weaker tool-use training.
Temperature is set to 0 where supported. Each agent--model combination is evaluated over all legs; we report per-difficulty and aggregate results.

\noindent\textbf{Agent interface.}
Each agent receives: (i)~the seed Wikipedia URL; (ii)~the clue-envelope text; (iii)~schema descriptions of all 19 tools; and (iv)~a step budget of $B = \max(10, \lfloor 1.5K \rfloor)$ where $K$ is the number of pit stops. The agent must produce a single digit 0--9 as its answer. Tool outputs longer than 8{,}000 characters are truncated.

\noindent\textbf{Uniform timeout.}
All agents receive a uniform wall-clock timeout of \textbf{600 seconds} per leg, regardless of difficulty level. We chose this budget based on analysis of completed trials: 92\% of correct answers on \aar{}-Linear and 95\% on \aar{}-DAG are produced within 600 seconds, while incorrect trials that run longer (up to 1{,}800s on extreme legs) overwhelmingly continue executing on wrong paths without recovering. A uniform timeout ensures fair cross-difficulty comparison and avoids inflating costs on legs where the agent is irretrievably lost. Each trial runs in a Docker container with 10{,}240\,MB memory and internet access for tool API calls.

\noindent\textbf{Metrics.}
We report three primary metrics and two supplementary indicators: (1) \textbf{Finish-line accuracy} (FA): Whether the agent's single-digit answer matches the golden finish-line code. This is the primary success metric, (2) \textbf{Pit-stop visit rate} (PVR): The fraction of golden \texttt{route\_info} pit stops for which the agent fetched the correct Wikipedia URL, measuring navigation quality, and (3) \textbf{Roadblock completion rate} (RCR): The fraction of golden \texttt{roadblock} pit stops for which the agent invoked all expected tools in the chain, measuring tool-use competence, as well as \textbf{Average steps}: Mean number of LLM turns per leg (lower is more efficient) and \textbf{Step-limit hit rate}: Fraction of legs where the agent exhausted its budget without producing an answer.

\noindent\textbf{Baselines.}
To calibrate our metrics, we include a \textbf{random} baseline that outputs a uniformly random digit 0--9 (expected FA = 10\%, PVR = 0\%, RCR = 0\%). This establishes the chance-level floor for the single-digit finish-line code.

\noindent\textbf{Cost and reproducibility.}
The full evaluation (7{,}000 trials across 10 configurations) consumed 286 compute-hours. Token usage varies by 10$\times$ across frameworks: Codex CLI averages 1.4--1.8M tokens/trial, while mini-swe-agent uses 149--187K. Claude Code achieves comparable accuracy to Codex CLI (37.2\% vs.\ 37.1\%) with 6$\times$ fewer tokens. All golden execution traces and Wikipedia snapshots are cached for deterministic re-scoring. All trials use temperature 0 for deterministic outputs; variance arises only from live API responses, which are cached at generation time. Full resource breakdown in Appendix~\ref{appendix:resources}.

\vspace{-3mm}
\section{Results}
\vspace{-3mm}
We evaluate \aar{} along three axes: (1)~how do current LLMs perform across difficulty levels? (2)~where in the navigation--tool--reasoning pipeline do agents fail? and (3)~how do different agent architectures compare?
\label{sec:experiments}

\definecolor{facolor}{HTML}{E6550D}
\definecolor{pvrcolor}{HTML}{2171B5}
\definecolor{rcrcolor}{HTML}{238B45}
 
\newcommand{\FA}{\textcolor{facolor}{\textbf{FA}}}
\newcommand{\PVR}{\textcolor{pvrcolor}{\textbf{PVR}}}
\newcommand{\RCR}{\textcolor{rcrcolor}{\textbf{RCR}}}

\subsection{Main Results}
\label{sec:main_results}
\begin{figure}[t]
\centering
\includegraphics[width=\linewidth]{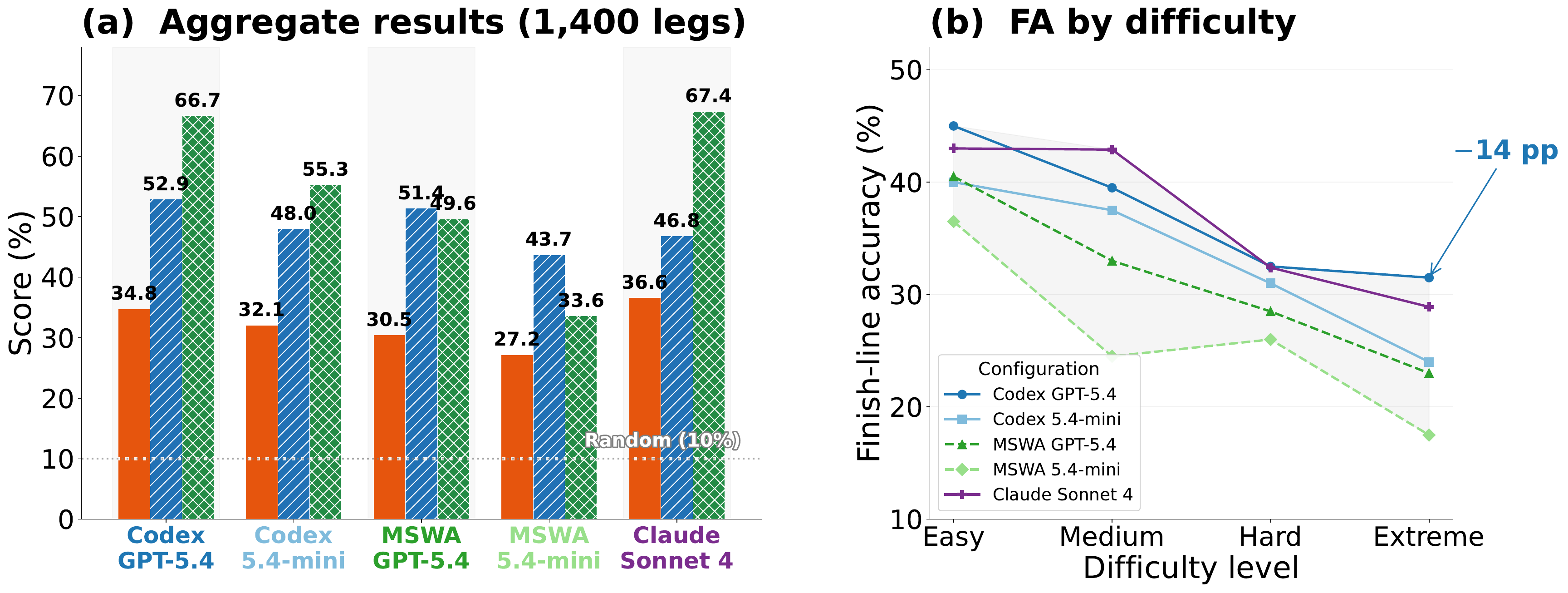}
\caption{
  \textbf{(a)} Aggregate results across all 1{,}400 legs (weighted average of Linear and DAG). \textcolor{facolor}{\textbf{FA}} (finish-line accuracy), \textcolor{pvrcolor}{\textbf{PVR}} (navigation), \textcolor{rcrcolor}{\textbf{RCR}} (tool use). Best FA is 36.6\% (Claude + Sonnet 4); \textcolor{pvrcolor}{PVR} is consistently the weakest metric.
  \textbf{(b)} FA degrades monotonically with difficulty (best: $-$13.5\,pp, worst: $-$19.0\,pp). Per-variant breakdown in Appendix~\ref{appendix:per_difficulty_figs}.
}
\label{fig:overview}
\vspace{-5mm}
\end{figure}

Figure~\ref{fig:overview} presents main results across both benchmark variants. No configuration exceeds 37.2\% FA, with PVR (navigation) consistently the weakest metric. Agent architecture matters as much as model scale: Codex + GPT-5.4 and Claude Code + Sonnet 4 tie at 37\% despite different providers, while the full spread across configs is 11pp. Full per-difficulty results are in Table~\ref{tab:main_results_appenidx} (Appendix).

\subsection{Key Findings}

\noindent\textbf{Finding 1: Difficulty degrades accuracy, driven by navigation.}
FA decreases with difficulty across all configs (Figure~\ref{fig:overview}b): Codex + GPT-5.4 drops from 45.0\% (easy) to 31.5\% (extreme), Claude Code from 43.0\% to 28.9\%. PVR drops sharply (88.7\% $\to$ 37.1\%) while RCR declines more gently (83.6\% $\to$ 49.2\%), confirming navigation as the primary difficulty driver.

\noindent\textbf{Finding 2: Navigation is the primary bottleneck, not tool use.}
Error decomposition (Table~\ref{tab:error_decomp}) confirms: navigation errors account for 30.9\% of all trials (rising to 52\% at extreme difficulty) versus only 8.6\% for tool-use errors. This pattern holds across all configurations.

\noindent\textbf{Finding 3: Agent architecture matters as much as model scale.}
The framework gap (Codex CLI vs.\ mini-swe-agent) is larger than the model-scale gap (GPT-5.4 vs.\ GPT-5.4-mini). Codex + GPT-5.4 (37.1\%) outperforms mini-swe + GPT-5.4-mini (26.1\%) by 11pp, while Claude Code + Sonnet 4 matches at 37.2\% despite a different provider. The key differentiator is tool-use competence: Codex CLI achieves 65.8\% RCR (tool use) vs.\ 34.4\% for mini-swe-agent. Mini-swe-agent under-explores (8 to 9 steps vs.\ 34 to 48 for Codex), committing to answers before sufficient verification. On \aar{}-DAG, Claude Code achieves the highest RCR (71.6\%), indicating strong compositional tool-use despite lower PVR. Notably, token efficiency varies by 10$\times$: Claude Code matches Codex CLI on accuracy (37.2\% vs.\ 37.1\%) while consuming 6$\times$ fewer tokens per trial (114--225K vs.\ 1.4--1.8M), suggesting that task performance and token usage are largely decoupled in current agent architectures.

\noindent\textbf{Finding 4: Reasoning models fail under time constraints.}
Codex CLI + GPT-OSS-120B (120B open-weight reasoning model) achieves only 3.1\% FA on \aar{}-Linear, barely above the 10\% random baseline. The model spends its budget on internal reasoning (2.2 tool calls vs.\ 27 for GPT-5.4), completing just $\sim$1 agent turn before timeout. Extended thinking is counterproductive for agentic tasks requiring many shallow tool calls (full analysis in Appendix~\ref{appendix:reasoning_models}).

\vspace{-3mm}
\subsection{Linear vs.\ Compositional: The Impact of DAG Structure}
\label{sec:compositional_results}
\vspace{-3mm}

\begin{wrapfigure}[12]{r}{0.6\linewidth}
  \vspace{-4mm}
  \centering
  \includegraphics[width=\linewidth]{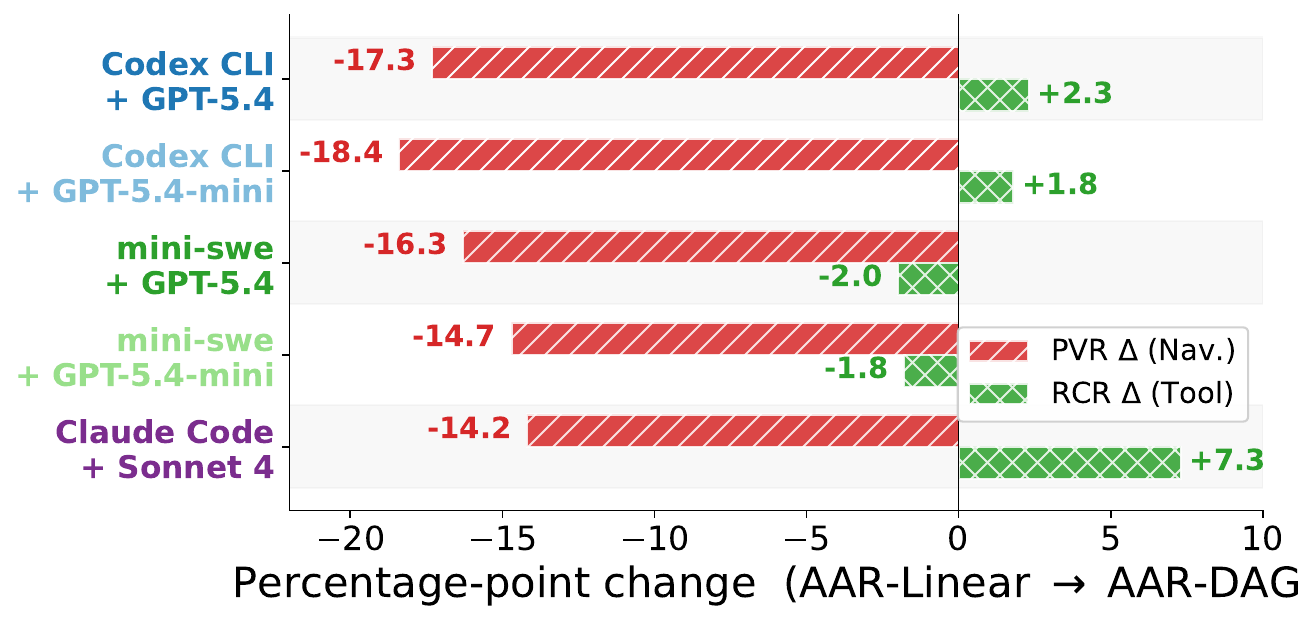}\vspace{-4mm}
  \caption{%
\textbf{DAG structure penalizes navigation, not tool use.}}
  \label{fig:linear-dag-delta}
\end{wrapfigure}

Having established baseline performance on \aar{}-Linear, we now examine how compositional DAG structure affects these results. Comparing the two variants (Figure~\ref{fig:overview}a) reveals a consistent pattern across all configurations.

\textbf{Finding 5: Compositionality penalizes navigation, not tool use.}
As shown in Figure~\ref{fig:linear-dag-delta}, PVR drops by 13--18\,pp
from \aar{}-Linear to \aar{}-DAG (agents visit fewer required Wikipedia
pages on longer trails), while RCR remains stable or even increases
slightly.
Finish-line accuracy drops modestly for stronger configurations
($-$5.5\,pp for Codex\,+\,GPT-5.4) but \emph{increases} for the weakest
($+$2.5\,pp for mini-swe-agent\,+\,GPT-5.4-mini).
This reinforces Finding~2: diamond fork--merge patterns do not confuse
agents who reach the right pages; the added difficulty comes entirely
from navigating longer trails.

\noindent\textbf{Finding 6: Shortcuts increase with compositionality.}
On \aar{}-DAG, 14--21\% of all trials achieve the correct answer while visiting $<$30\% of required pages (vs.\ 6--11\% on \aar{}-Linear). Shortcuts are not lucky guesses (43.8\% RCR, 60.9\% intermediate accuracy) but reflect agents inferring tool arguments from clue context. Our decomposed metrics explicitly detect this: PVR\,$<$\,0.3 flags navigation bypass. Detailed shortcut analysis is in Appendix~\ref{appendix:shortcuts}.

\vspace{-3mm}
\subsection{Error Decomposition}
\label{sec:analysis}
\vspace{-3mm}

\begin{wraptable}[15]{r}{0.45\textwidth}
\centering
\small\vspace{-10mm}
\setlength{\tabcolsep}{3pt}
\begin{tabular}{@{}ll cccc@{}}
\toprule
& \textbf{Level} & \textbf{Nav.} & \textbf{Tool} & \textbf{Comp.} & \textbf{Corr.} \\
\midrule
\multirow{5}{*}{\rotatebox[origin=c]{90}{\scriptsize Linear}}
& Easy    & 5.0 & 15.0 & 40.0 & 40.0 \\
& Medium  & 20.0 & 10.0 & 32.5 & 37.5 \\
& Hard    & 46.5 & 6.0  & 16.5 & 31.0 \\
& Extreme & 52.0 & 3.5  & 20.5 & 24.0 \\
& \cellcolor{gray!10}\textit{All} & \cellcolor{gray!10}\textit{30.9} & \cellcolor{gray!10}\textit{8.6} & \cellcolor{gray!10}\textit{27.4} & \cellcolor{gray!10}\textit{33.1} \\
\midrule
\multirow{5}{*}{\rotatebox[origin=c]{90}{\scriptsize DAG}}
& Easy    & 24.0 & 9.0  & 41.0 & 26.0 \\
& Medium  & 35.3 & 6.0  & 28.7 & 30.0 \\
& Hard    & 59.0 & 2.4  &  9.0 & 29.5 \\
& Extreme & 59.2 & 0.5  &  5.4 & 34.8 \\
& \cellcolor{gray!10}\textit{All} & \cellcolor{gray!10}\textit{47.3} & \cellcolor{gray!10}\textit{3.8} & \cellcolor{gray!10}\textit{18.2} & \cellcolor{gray!10}\textit{30.7} \\
\bottomrule
\end{tabular}
\caption{Error decomposition (\%) for Codex CLI + GPT-5.4-mini. Nav. errors increase +16pp on DAG while tool errors \emph{decrease} $-$5pp despite 3$\times$ longer chains.}
\label{tab:error_decomp}
\end{wraptable}
Table~\ref{tab:error_decomp} decomposes trials into navigation (PVR $< 0.5$), tool-use (PVR $\geq 0.5$, RCR $< 0.5$), and computation errors (both $\geq 0.5$, FA $= 0$). Navigation errors grow from 5\% (easy) to 52\% (extreme); computation errors peak on easy legs (40\%); tool-use errors remain moderate.
On \aar{}-DAG, navigation errors increase to 47.3\% (+16pp) while tool-use errors \emph{decrease} to 3.8\% ($-$5pp) despite 3$\times$ longer chains, suggesting diamond riddles provide clearer tool-invocation cues.

Additional analyses in the appendix cover per-template tool-use patterns (Appendix~\ref{appendix:template_analysis}), scaling behavior by leg length (Appendix~\ref{appendix:scaling}), and recovery rates from partial success (Appendix~\ref{appendix:recovery}).

\vspace{-3mm}
\subsection{Discussion}
\label{sec:discussion}
\vspace{-3mm}

Manual inspection of 50 failed legs reveals five failure modes: planning without execution (fabricated observations), argument mis-routing between tool-chain steps, arithmetic errors in finish-line computation, navigation drift on longer legs, and step budget exhaustion. As a concrete example, on an extreme-difficulty leg (36 stops), Codex + GPT-5.4-mini visits only 1 of 14 required pages (PVR\,=\,0.07) yet invokes every expected tool type (RCR\,=\,1.0), applying them to \emph{wrong} pages. The agent self-corrects between wrong candidates, demonstrating that iterative hypothesis refinement amplifies errors when initial navigation is off. A single accuracy score hides this: decomposed metrics reveal perfect tool competence with failed navigation. Additional case studies are in Appendix~\ref{appendix:case_studies}.

The step budget is sufficient (hit rate $<$1.5\% across configs), and metric decomposition confirms PVR and RCR capture distinct failure modes: navigation-only failures are common while tool-only failures are rare. Fine-grained analysis (Appendix~\ref{appendix:discussion}) reveals 20.5\% of trials are \emph{near-misses} ($\geq$80\% intermediate accuracy, wrong final answer), and incorrect trials paradoxically make \emph{more} tool calls than correct ones (21.7 vs.\ 16.5), indicating over-exploration on wrong pages. These findings suggest that improving \emph{targeted retrieval}, not increasing search volume, is the key opportunity: incorrect trials already issue 56\% more searches and fetch 18\% more pages than correct ones. Full analysis in Appendix~\ref{appendix:discussion}.

\vspace{-3mm}
\section{Conclusion}
\label{sec:conclusion}
\vspace{-3mm}

We presented \aar{}, a DAG-structured benchmark with three decomposed metrics (FA, PVR, RCR) that separately diagnose navigation, tool-use, and computation failures. Across 1{,}400 legs and three agent frameworks, the best achieves only 37.2\% FA: agents are competent tool users but poor navigators, and compositional structure amplifies this gap.

\aar{} uses Wikipedia as its sole navigation source with 19 tools. We plan to expand to broader domains (calendars, databases), introduce richer DAG topologies (shared sub-expressions, conditional branches), support multi-leg seasons with cross-episode state, and develop partial-credit evaluation via calibrated LLM judges.


\section*{Acknowledgments}
We thank members of Minnesota NLP for their insightful input during group meetings. We also extend our appreciation to Chanwoo Park and Yuxin Chen for their initial research contribution and discussion. ZMK is generously supported by the 3M Science and Technology Fellowship and the Doctoral Dissertation Fellowship at the University of Minnesota.

\section*{Ethics Statement}

\aar{} uses publicly available Wikipedia content under the Creative Commons Attribution-ShareAlike License and queries commercial APIs (Google Maps, Yahoo Finance, Binance, Serper) within their terms of service. We do not collect, store, or redistribute personal data. Our Wikipedia crawler respects \texttt{robots.txt} and rate limits. The benchmark does not involve human subjects. We acknowledge the environmental cost of running large-scale LLM evaluations and mitigate this by caching golden execution traces for deterministic re-scoring without repeated API calls. The benchmark is intended for research evaluation of agent capabilities and should not be used to make deployment decisions without additional domain-specific validation. The code and data can be accessed at: 

\section*{Reproducibility Statement}

All \aar{} instances are deterministically reproducible: each leg includes cached Wikipedia page snapshots, golden execution traces with all intermediate values, and the finish-line code $y^*$, enabling re-scoring independent of live API state. The generation pipeline uses GPT-4o for route planning and clue verbalization, with temperature 0 for determinism. Tool chains are executed against live APIs at generation time, and their outputs are cached alongside each leg. The evaluation framework specifies: model temperature 0, step budget formula $B = \max(10, \lfloor 1.5K \rfloor)$, tool output truncation at 8{,}000 characters, and 19 tool schemas provided to each agent. Code for generation and evaluation, the full dataset, and all experimental scripts are available at \url{https://github.com/minnesotanlp/the-amazing-agent-race} under the MIT License. The dataset will be hosted on HuggingFace upon acceptance with a datasheet documenting data collection, annotation, and intended use per~\citet{gebru2021datasheets}.

\bibliography{colm2026_conference}

@misc{amazingrace,
  title={The Amazing Race},
  author={{CBS}},
  year={2001--present},
  note={American reality television series created by Elise Doganieri and Bertram van Munster},
  howpublished={\url{https://en.wikipedia.org/wiki/The_Amazing_Race_(American_TV_series)}}
}

@article{hendrycks2021mmlu,
  title={Measuring massive multitask language understanding},
  author={Hendrycks, Dan and Burns, Collin and Basart, Steven and Zou, Andy and Mazeika, Mantas and Song, Dawn and Steinhardt, Jacob},
  journal={arXiv preprint arXiv:2009.03300},
  year={2021}
}

@article{cobbe2021gsm8k,
  title={Training verifiers to solve math word problems},
  author={Cobbe, Karl and Kosaraju, Vineet and Bavarian, Mohammad and Chen, Mark and Jun, Heewoo and Kaiser, Lukasz and Plappert, Matthias and Tworek, Jerry and Hilton, Jacob and Nakano, Reiichiro and others},
  journal={arXiv preprint arXiv:2110.14168},
  year={2021}
}

@article{chen2021humaneval,
  title={Evaluating large language models trained on code},
  author={Chen, Mark and Tworek, Jerry and Jun, Heewoo and Yuan, Qiming and Pinto, Henrique Ponde de Oliveira and Kaplan, Jared and Edwards, Harri and Burda, Yuri and Joseph, Nicholas and Brockman, Greg and others},
  journal={arXiv preprint arXiv:2107.03374},
  year={2021}
}

@inproceedings{qin2024toolllm,
  title={{ToolLLM}: Facilitating large language models to master 16000+ real-world {APIs}},
  author={Qin, Yujia and Liang, Shihao and Ye, Yining and Zhu, Kunlun and Yan, Lan and Lu, Yaxi and Lin, Yankai and Cong, Xin and Tang, Xiangru and Qian, Bill and others},
  booktitle={International Conference on Learning Representations},
  year={2024}
}

@article{guo2024stabletoolbench,
  title={{StableToolBench}: Towards stable large-scale benchmarking on tool learning of large language models},
  author={Guo, Zhicheng and Cheng, Sijie and Wang, Hao and Liang, Shihao and Qin, Yujia and Li, Peng and Liu, Zhiyuan and Sun, Maosong and Liu, Yang},
  journal={Findings of the Association for Computational Linguistics: ACL 2024},
  year={2024}
}

@inproceedings{patil2025bfcl,
  title={The {B}erkeley Function Calling Leaderboard: From tool use to agentic evaluation of large language models},
  author={Patil, Shishir and Zhang, Tianjun and Call, Xin and others},
  booktitle={International Conference on Machine Learning},
  year={2025}
}

@inproceedings{li2023apibank,
  title={{API-Bank}: A comprehensive benchmark for tool-augmented {LLMs}},
  author={Li, Minghao and Song, Feifan and Yu, Bowen and Yu, Haiyang and Li, Zhoujun and Huang, Fei and Li, Yongbin},
  booktitle={Proceedings of the 2023 Conference on Empirical Methods in Natural Language Processing},
  year={2023}
}

@inproceedings{chen2024teval,
  title={{T-Eval}: Evaluating the tool utilization capability of large language models step by step},
  author={Chen, Zehui and Du, Weihua and Zhang, Wenwei and Liu, Kuikun and Liu, Jiangning and Zheng, Miao and Zhuo, Jingming and Zhang, Songyang and Lin, Dahua and Chen, Kai and Zhao, Feng},
  booktitle={Proceedings of the 62nd Annual Meeting of the Association for Computational Linguistics},
  year={2024}
}

@inproceedings{shen2024taskbench,
  title={{TaskBench}: Benchmarking large language models for task automation},
  author={Shen, Yongliang and Song, Kaitao and Tan, Xu and Zhang, Wenqi and others},
  booktitle={Advances in Neural Information Processing Systems},
  year={2024}
}

@article{basu2024nestful,
  title={{NESTFUL}: A benchmark for evaluating {LLMs} on nested sequences of {API} calls},
  author={Basu, Kinjal and Abdelaziz, Ibrahim and Kate, Kiran and Agarwal, Mayank and Crouse, Maxwell and Rizk, Yara and others},
  journal={arXiv preprint arXiv:2409.03797},
  year={2024}
}

@article{ye2025toolhop,
  title={{ToolHop}: A query-driven benchmark for evaluating large language models in multi-hop tool use},
  author={Ye, Jiarui and others},
  journal={Proceedings of the 63rd Annual Meeting of the Association for Computational Linguistics},
  year={2025}
}

@article{lu2025toolsandbox,
  title={{ToolSandbox}: A stateful, conversational, interactive evaluation benchmark for {LLM} tool use capabilities},
  author={Lu, Jiarui and others},
  journal={Findings of the North American Chapter of the Association for Computational Linguistics},
  year={2025}
}

@inproceedings{zhou2024webarena,
  title={{WebArena}: A realistic web environment for building autonomous agents},
  author={Zhou, Shuyan and Xu, Frank F and Zhu, Hao and Zhou, Xuhui and Lo, Robert and Sridhar, Abishek and Cheng, Xianyi and Ou, Tianyue and Bisk, Yonatan and Fried, Daniel and others},
  booktitle={International Conference on Learning Representations},
  year={2024}
}

@inproceedings{deng2024mind2web,
  title={{Mind2Web}: Towards a generalist agent for the web},
  author={Deng, Xiang and Gu, Yu and Zheng, Boyuan and Chen, Shijie and Stevens, Sam and Wang, Boshi and Sun, Huan and Su, Yu},
  booktitle={Advances in Neural Information Processing Systems},
  year={2024}
}

@inproceedings{xie2024osworld,
  title={{OSWorld}: Benchmarking multimodal agents for open-ended tasks in real computer environments},
  author={Xie, Tianbao and Zhang, Danyang and Chen, Jixuan and Li, Xiaochuan and others},
  booktitle={Advances in Neural Information Processing Systems},
  year={2024}
}

@inproceedings{liu2023agentbench,
  title={{AgentBench}: Evaluating {LLMs} as agents},
  author={Liu, Xiao and Yu, Hao and Zhang, Hanchen and Xu, Yifan and Lei, Xuanyu and Lai, Hanyu and Gu, Yu and Ding, Hangliang and Men, Kaiwen and Yang, Kejuan and others},
  booktitle={International Conference on Learning Representations},
  year={2024}
}

@inproceedings{mialon2023gaia,
  title={{GAIA}: A benchmark for general {AI} assistants},
  author={Mialon, Gr{\'e}goire and Fourrier, Cl{\'e}mentine and Swift, Craig and Wolf, Thomas and LeCun, Yann and Scialom, Thomas},
  booktitle={International Conference on Learning Representations},
  year={2024}
}

@inproceedings{ma2024agentboard,
  title={{AgentBoard}: An analytical evaluation board of multi-turn {LLM} agents},
  author={Ma, Chang and Zhang, Junlei and Zhu, Zhihao and Yang, Cheng and others},
  booktitle={Advances in Neural Information Processing Systems},
  year={2024}
}

@inproceedings{trivedi2024appworld,
  title={{AppWorld}: A controllable world of apps and people for benchmarking interactive coding agents},
  author={Trivedi, Harsh and Khot, Tushar and Hartmann, Mareike and Manku, Ruskin and others},
  booktitle={Proceedings of the 62nd Annual Meeting of the Association for Computational Linguistics},
  year={2024}
}

@article{yao2024taubench,
  title={tau-bench: A benchmark for tool-agent-user interaction in real-world domains},
  author={Yao, Shunyu and Shinn, Noah and Razavi, Pedram and Narasimhan, Karthik},
  journal={arXiv preprint arXiv:2406.12045},
  year={2024}
}

@article{xu2024theagentcompany,
  title={{TheAgentCompany}: Benchmarking {LLM} agents on consequential real world tasks},
  author={Xu, Frank F and Song, Yufan and Li, Boxuan and others},
  journal={arXiv preprint arXiv:2412.14161},
  year={2024}
}

@article{mcpbench2025,
  title={{MCP-Bench}: Benchmarking tool-using {LLM} agents with complex real-world tasks via {MCP} servers},
  author={{Accenture Labs}},
  journal={arXiv preprint arXiv:2508.20453},
  year={2025}
}

@misc{harbor2026,
  author = {{Harbor Framework Team}},
  month = jan,
  title = {{Harbor: A framework for evaluating and optimizing agents and models in container environments}},
  url = {https://github.com/laude-institute/harbor},
  year = {2026}
}

@article{gebru2021datasheets,
  title={Datasheets for datasets},
  author={Gebru, Timnit and Morgenstern, Jamie and Vecchione, Briana and Vaughan, Jennifer Wortman and Wallach, Hanna and Iii, Hal Daum{\'e} and Crawford, Kate},
  journal={Communications of the ACM},
  volume={64},
  number={12},
  pages={86--92},
  year={2021}
}
\bibliographystyle{colm2026_conference}

\appendix

\section{Additional Related Work}
\label{appendix:more_related_work}

\noindent\textbf{Holistic agent benchmarks.}
AgentBench~\citep{liu2023agentbench} spans eight environments from OS interaction
to web shopping. AgentBoard~\citep{ma2024agentboard} adds a Progress Rate metric
for richer subgoal signal. AppWorld~\citep{trivedi2024appworld} evaluates coding
agents across 457 APIs in nine simulated apps. tau-bench~\citep{yao2024taubench}
targets tool-agent-user interaction (GPT-4o: $<$50\% pass\textsuperscript{1}).
TheAgentCompany~\citep{xu2024theagentcompany} benchmarks professional tasks with
checkpoint-based partial credit (best model: 30\%). These benchmarks trade depth
for breadth; \aar{} makes the complementary trade-off, probing the
navigation--tool--reasoning pipeline with structurally controlled difficulty and
three metrics that independently diagnose each failure stage.

\noindent\textbf{Contamination resistance.}
Fixed benchmarks such as MMLU~\citep{hendrycks2021mmlu}, GSM8K~\citep{cobbe2021gsm8k},
and HumanEval~\citep{chen2021humaneval} face growing contamination as instances
appear in training corpora. MCP-Bench~\citep{mcpbench2025} uses live MCP servers
(250 tools, 28 servers) but relies on manual curation. \aar{} seeds each instance
from a random Wikipedia article and touches live APIs (stock prices, cryptocurrency
volumes, weather) that change daily; clue paraphrasing, analytical transforms, and
multi-step aggregation ensure answers cannot be recalled from training data
(\S\ref{sec:quality}).

\section{Difficulty Level Parameters}
\label{appendix:difficulty}

\begin{table}[H]
\centering
\small
\begin{tabular}{@{}lccccccc@{}}
\toprule
\textbf{Level} & \textbf{Pit Stops} & \textbf{Roadblocks} & \textbf{Detours} & \textbf{Diamonds} & \textbf{Extraction} & \textbf{Crawl} \\
\midrule
Easy    & 3--6   & 1--2 & 1--2 & 1   & infobox, prose & 1 \\
Medium  & 7--12  & 2--4 & 2--3 & 1--2 & + cross-section & 2 \\
Hard    & 13--16 & 4--5 & 3--4 & 2--3 & + cross-section & 3 \\
Extreme & 17--21 & 5--7 & 4--6 & 3--5 & + cross-section & 3 \\
\bottomrule
\end{tabular}
\caption{Difficulty level parameters (pre-augmentation). \textbf{Pit Stops}: configured range before diamond insertion. \textbf{Diamonds}: fork--merge patterns that create non-linear DAG dependencies (\S\ref{sec:diamonds}). \textbf{Crawl}: Wikipedia link-graph hops available for route planning. After diamond augmentation, each diamond adds 3 stops (two branches + merge), so actual pit-stop counts exceed these ranges.}
\label{tab:difficulty}
\end{table}

\section{Benchmark Validity}
\label{appendix:validity}

\paragraph{Gold plan solvability.}
By construction, every leg in the evaluation set has been solved by the golden executor, confirming that each instance is solvable with the provided tool set. We additionally verify that the clue envelope unambiguously implies the golden answer via round-trip validation (\S\ref{sec:quality}).

\paragraph{Inter-instance diversity.}
We measure diversity by computing pairwise Jaccard similarity between the sets of Wikipedia pages visited across all 800 legs (10{,}000 random pairs sampled).
The mean Jaccard similarity is \textbf{0.0005}, with 99.1\% of pairs sharing \emph{zero} pages.
This near-zero overlap confirms that random Wikipedia seeding produces highly diverse instances with negligible content overlap, making memorization-based shortcuts ineffective.

\paragraph{Temporal stability.}
Because some tools return live data (weather, elevation), temporal stability is an important concern. By design, \aar{} mitigates this through two mechanisms: (1)~golden execution traces are cached at generation time, so re-scoring uses deterministic reference values regardless of current API state; and (2)~the finish-line computation uses modular arithmetic ($\texttt{mod10}$, $\texttt{digital\_root}$), which absorbs small perturbations in tool outputs. Moreover, 15 of the 17 roadblock templates query temporally stable data (elevation, coordinates, country statistics, place counts), while only stock and crypto templates depend on date-specific market data that is fixed at generation time.

\section{Tool Set and Roadblock Templates}
\label{appendix:templates}

Table~\ref{tab:tools_full} lists the 19 tools available to agents, and Table~\ref{tab:templates} lists the 17 roadblock templates.

\paragraph{Argument passing.}
Tool-chain pit stops are instantiated from 17 predefined templates that compose 1--3 tools. Arguments flow between steps via three special keys: \texttt{\_\_from\_previous} (merge the output dict), \texttt{\_\_from\_previous\_as\_locations} (wrap coordinates for elevation), and \texttt{\_\_from\_previous\_as\_origins\_destinations} (format for the distance matrix).

\begin{table}[H]
\centering
\small
\begin{tabular}{@{}llp{4.5cm}@{}}
\toprule
\textbf{Category} & \textbf{Tool} & \textbf{Description} \\
\midrule
\multirow{2}{*}{Fetch \& Search} & \texttt{fetch\_webpage} & Fetch and parse web content \\
& \texttt{web\_search} & Google search via Serper API \\
\midrule
\multirow{7}{*}{Google Maps} & \texttt{maps\_geocode} & Address to coordinates \\
& \texttt{maps\_reverse\_geocode} & Coordinates to address \\
& \texttt{maps\_search\_places} & Search nearby places \\
& \texttt{maps\_place\_details} & Place metadata and ratings \\
& \texttt{maps\_distance\_matrix} & Driving distances \\
& \texttt{maps\_elevation} & Elevation at coordinates \\
& \texttt{maps\_directions} & Directions and duration \\
\midrule
\multirow{2}{*}{Weather} & \texttt{weather\_historical} & Historical weather data \\
& \texttt{weather\_forecast} & Weather forecasts \\
\midrule
\multirow{2}{*}{Code} & \texttt{python\_execute\_code} & Run Python code \\
& \texttt{python\_generate\_code} & LLM-generated Python \\
\midrule
\multirow{2}{*}{Countries} & \texttt{countries\_population} & Population data \\
& \texttt{countries\_area} & Area in km\textsuperscript{2} \\
\midrule
\multirow{2}{*}{Stocks} & \texttt{stock\_historical\_price} & Closing price on a date \\
& \texttt{stock\_volume} & Trading volume on a date \\
\midrule
\multirow{2}{*}{Crypto} & \texttt{crypto\_historical\_price} & Crypto closing price on a date \\
& \texttt{crypto\_volume} & 24h trading volume on a date \\
\bottomrule
\end{tabular}
\caption{The 19 tools available to agents, organized by category.}
\label{tab:tools_full}
\end{table}

\begin{table}[H]
\centering
\small
\begin{tabular}{@{}lll@{}}
\toprule
\textbf{Template} & \textbf{Requires} & \textbf{Produces} \\
\midrule
geocode\_elevation & location & elevation \\
geocode\_weather\_historical & location, date & temperature \\
geocode\_weather\_precipitation & location, date & precipitation \\
geocode\_distance & 2 locations & distance \\
geocode\_directions\_duration & 2 locations & duration \\
date\_computation & date & day count \\
math\_conversion & numeric value & converted value \\
nearby\_poi\_count & location & POI count \\
place\_rating & location & rating \\
country\_population & country & population \\
country\_area & country & area (km$^2$) \\
historical\_snowfall & location, date & snowfall \\
historical\_sunshine & location, date & sunshine hours \\
\midrule
stock\_price & ticker, date & closing price \\
stock\_volume & ticker, date & trading volume \\
crypto\_price & crypto pair, date & closing price \\
crypto\_volume & crypto pair, date & 24h volume \\
\bottomrule
\end{tabular}
\caption{Roadblock templates. Each composes 1--3 tool calls.}
\label{tab:templates}
\end{table}

\section{Per-Template Tool-Use Analysis}
\label{appendix:template_analysis}

Table~\ref{tab:template_rates} shows finish-line accuracy broken down by which roadblock template appears in a leg. Pure computation templates (\texttt{date\_computation}: 40.2\%, \texttt{math\_conversion}: 33.4\%) are easiest---agents execute Python code reliably once they have input values. Geographic API templates (\texttt{geocode\_elevation}: 27.0\%, \texttt{nearby\_poi}: 28.1\%) fall in the middle. The hardest templates involve specialized APIs: \texttt{stock\_price} (18.5\%), \texttt{weather} (22.2\%), and \texttt{place\_rating} (22.5\%), which require precise parameter formatting that agents frequently get wrong. The pattern is consistent across all four configurations.

\begin{table}[H]
\centering
\small
\begin{tabular}{@{}lr cccc c@{}}
\toprule
& & \multicolumn{4}{c}{\textbf{FA (\%) by Config}} & \\
\cmidrule(lr){3-6}
\textbf{Template} & \textbf{N} & \textbf{C\,5.4} & \textbf{C\,m} & \textbf{M\,5.4} & \textbf{M\,m} & \textbf{Avg} \\
\midrule
date\_comp       & 202 & 48.0 & 40.6 & 40.6 & 31.7 & 40.2 \\
math\_conv       & 298 & 39.9 & 35.2 & 30.2 & 28.2 & 33.4 \\
geocode\_dist    &  41 & 39.0 & 34.1 & 36.6 & 22.0 & 32.9 \\
country\_area    &  44 & 31.8 & 29.5 & 31.8 & 22.7 & 29.0 \\
country\_pop     & 209 & 34.4 & 29.2 & 28.2 & 22.0 & 28.5 \\
nearby\_poi      & 506 & 32.6 & 29.8 & 29.1 & 20.9 & 28.1 \\
geocode\_elev    & 392 & 32.9 & 27.8 & 25.8 & 21.7 & 27.0 \\
directions       &  60 & 35.0 & 26.7 & 20.0 & 23.3 & 26.2 \\
stock\_vol       &  26 & 34.6 & 19.2 & 30.8 & 19.2 & 26.0 \\
place\_rating    & 162 & 29.6 & 21.0 & 22.8 & 16.7 & 22.5 \\
weather          &  25 & 33.3 & 25.9 & 22.2 &  7.4 & 22.2 \\
stock\_price     &  54 & 25.9 & 18.5 & 16.7 & 13.0 & 18.5 \\
\bottomrule
\end{tabular}
\caption{Per-template FA (\%) on \aar{}-Linear. \textbf{N}: legs containing this template. \textbf{C}: Codex CLI. \textbf{M}: mini-swe-agent. \textbf{m}: GPT-5.4-mini.}
\label{tab:template_rates}
\end{table}

\section{Scaling Behavior}
\label{appendix:scaling}

Table~\ref{tab:scaling} shows how metrics scale with leg length for Codex CLI + GPT-5.4-mini. PVR declines sharply from 83.5\% (3--8 stops) to 35.8\% (27--40 stops), while RCR declines from 71.6\% to 37.5\%. Finish-line accuracy also declines steadily from 40.2\% (short legs) to 17.4\% (long legs), confirming that longer chains compound navigation errors into lower overall accuracy.

\begin{table}[H]
\centering
\small
\begin{tabular}{@{}lcccr@{}}
\toprule
\textbf{Stops} & \textbf{FA (\%)} & \textbf{PVR (\%)} & \textbf{RCR (\%)} & \textbf{N} \\
\midrule
3--8    & 40.2 & 83.5 & 71.6 & 199 \\
9--14   & 36.2 & 62.4 & 57.5 & 185 \\
15--20  & 32.6 & 41.3 & 47.1 & 215 \\
21--26  & 24.7 & 38.6 & 44.3 & 178 \\
27--40  & 17.4 & 35.8 & 37.5 &  23 \\
\bottomrule
\end{tabular}
\caption{Scaling behavior: FA, PVR, and RCR as a function of leg length (number of pit stops) for Codex CLI + GPT-5.4-mini on \aar{}-Linear.}
\label{tab:scaling}
\end{table}

\section{Agent Recovery from Partial Success}
\label{appendix:recovery}

We analyze how effectively agents convert partial success into correct final answers. Table~\ref{tab:recovery} summarizes recovery rates: FA conditioned on high partial metrics.

On \aar{}-Linear, Codex + GPT-5.4 converts high-PVR ($\geq 0.8$) legs to correct answers 45.0\% of the time (282 legs). When \emph{both} PVR and RCR are high, recovery rises to 50.3\% (199 legs)---confirming that getting both navigation and tool use right is necessary but not sufficient.
On \aar{}-DAG, recovery rates drop notably: Codex + GPT-5.4 converts both-high legs at only 31.7\% (60 legs).
This 19pp drop reveals that compositional finish-line expressions (aggregating values through diamond merge points) are substantially harder to compute correctly, even when all inputs are available. Across the board, the linear-to-compositional transition reduces recovery by 10--19pp.

\begin{table}[H]
\centering
\small
\begin{tabular}{@{}l cc cc@{}}
\toprule
& \multicolumn{2}{c}{\textbf{PVR $\geq$ 0.8 $\to$ FA\,=\,1}} & \multicolumn{2}{c}{\textbf{Both $\geq$ 0.8 $\to$ FA\,=\,1}} \\
\cmidrule(lr){2-3} \cmidrule(lr){4-5}
\textbf{Config} & \textbf{Lin.} & \textbf{DAG} & \textbf{Lin.} & \textbf{DAG} \\
\midrule
Codex + 5.4       & 45.0 & 30.6 & 50.3 & 31.7 \\
Codex + 5.4-mini  & 44.1 & 36.0 & 47.0 & 30.0 \\
MSWA + 5.4        & 43.6 & 30.8 & 48.3 & 39.4 \\
MSWA + 5.4-mini   & 38.5 & 28.9 & 39.0 & 62.5$^\dagger$ \\
\bottomrule
\end{tabular}
\caption{Recovery rates (\%): FA conditioned on high partial metrics. \textbf{Lin.}: \aar{}-Linear. \textbf{DAG}: \aar{}-DAG. $^\dagger$Based on only 8 legs.}
\label{tab:recovery}
\end{table}

\section{Discussion: What \aar{} Reveals About Agent Limitations}
\label{appendix:discussion}

\paragraph{Failure taxonomy.}
Fine-grained analysis of Codex CLI + GPT-5.4-mini on \aar{}-Linear reveals four distinct failure populations:

\begin{enumerate}[nosep]
    \item \textbf{Near-misses} (20.5\% of all trials): The agent achieves $\geq$80\% intermediate value accuracy but produces the wrong finish-line code. These legs have strong PVR (63.5\%) and RCR (71.4\%), indicating the agent was on the right track but made a computational error in the final aggregation.

    \item \textbf{Perfect-navigation failures} (12.8\%): The agent visits $\geq$90\% of required pages but still gets the wrong answer, with RCR at 69.2\%. These represent tool-chain or computation errors downstream of successful navigation.

    \item \textbf{Navigation-bypass successes} (7.4\%): Agents that get the correct answer despite visiting $<$30\% of required pages. These skew toward harder legs (25 hard, 21 extreme), suggesting that experienced tool reasoning can sometimes compensate for navigation failure.

    \item \textbf{Total failures} (8.9\% for Codex, 17.6\% for mini-swe-agent): Both PVR and RCR below 30\%. Mini-swe-agent's higher rate (2$\times$) reflects its under-exploration strategy.
\end{enumerate}

\paragraph{The over-calling paradox.}
Counter-intuitively, \emph{incorrect} trials use more tool calls on average (21.7) than correct trials (16.5) for Codex + GPT-5.4-mini. Agents that fail tend to over-explore rather than under-explore---they call tools on wrong pages, get confusing results, and spiral into increasingly misguided attempts. Tool-call validity is high ($>$98\%) in both cases, meaning agents rarely produce malformed calls. The problem is not \emph{how} they call tools but \emph{which} tools they call and \emph{on what data}.

\paragraph{Implications for agent design.}
Our results suggest three concrete directions:

\begin{itemize}[nosep]
    \item \textbf{Invest in targeted retrieval, not more searching.} Incorrect trials issue 56\% more web searches (8.1 vs.\ 5.2 per trial) and fetch 18\% more pages (9.2 vs.\ 7.8) than correct trials. The key improvements are query decomposition, relevance verification, and early backtracking.

    \item \textbf{Add arithmetic verification.} The 20.5\% near-miss rate shows many agents get almost everything right but fumble the final computation. On \aar{}-DAG, recovery rates drop by 19pp (Table~\ref{tab:recovery}), indicating compositional merge expressions are especially error-prone.

    \item \textbf{Calibrate exploration depth.} Mini-swe-agent's 9-step average produces 17.6\% total failures versus Codex's 8.9\% at 35 steps. However, Codex's incorrect trials \emph{over}-call (21.7 tools). Adaptive step budgets that scale with intermediate confidence could help.
\end{itemize}

\section{Additional Case Studies}
\label{appendix:case_studies}

\noindent\textbf{Case A: Computation error despite perfect navigation} (\texttt{easy-sample\_063}; 8 stops, FA\,=\,0, PVR\,=\,1.00, RCR\,=\,1.00).
The agent visits every required page and invokes every tool correctly, achieving 88\% intermediate value accuracy, yet produces the wrong finish-line code.

\begin{errorbox}[Agent Trajectory (abridged)]
\small
\textbf{Step 7:} ``I have two plausible interpretations for the Egypt clue, so I'm checking the actual Wikimedia page behind the search hit.'' \\[2pt]
\textbf{Step 8:} ``The Egypt clue now matches the 39-meter Nazlet El-Semman elevation from search results. I'm doing the final arithmetic.'' \\[2pt]
\textbf{Final:} Wrote \texttt{3} to answer.txt. \hfill \textcolor{errorframe}{\textbf{Expected: 5}}
\end{errorbox}

\noindent
The agent extracted the right values but misrouted them through the diamond merge and finish-line expression, producing 3 instead of 5.
This isolates \emph{compositional arithmetic} (aggregating values through fork--merge structures) as a distinct failure mode.

\vspace{0.5em}
\noindent\textbf{Case B: Correct extreme via tool-use shortcut} (\texttt{extreme-sample\_022}; 35 stops, FA\,=\,1, PVR\,=\,0.09, RCR\,=\,0.80).
The agent solves a 35-stop extreme leg correctly while visiting only $\sim$1 of 11 required Wikipedia pages.

\begin{successbox}[Agent Trajectory (abridged)]
\small
\textbf{Step 1:} ``I'm starting from the seed page and extracting each clue in sequence, then I'll use location and map tools only where the riddle needs external data.'' \\[2pt]
\textbf{Steps 2--9:} The agent calls \texttt{maps\_geocode}, \texttt{maps\_elevation}, \texttt{maps\_search\_places}, and \texttt{python\_execute\_code} 23 times, correctly computing 80\% of intermediate values without visiting the expected Wikipedia pages. \\[2pt]
\textbf{Final:} Wrote \texttt{1} to answer.txt. \hfill \textcolor{successframe}{\textbf{Correct!}}
\end{successbox}

\noindent
Rather than following the clue envelope's intended navigation path, the agent bypasses Wikipedia entirely and reasons directly about the tool outputs, achieving the correct answer through a ``tool-use shortcut.''

\section{Reasoning Model Analysis}
\label{appendix:reasoning_models}

Codex CLI + GPT-OSS-120B, an open-weight reasoning model with extended thinking, achieves only 3.1\% FA on \aar{}-Linear (clean trials), barely above the 10\% random baseline and 12$\times$ below GPT-5.4 (37.1\%).
The failure is not due to model size (120B parameters) but to \emph{execution strategy}: GPT-OSS-120B spends most of its token budget on internal reasoning, averaging only 2.2 tool calls per trial (vs.\ 27 for GPT-5.4) and completing just $\sim$1 agent turn before the 600s timeout.
Only 5\% of clean trials even write an answer.
On \aar{}-DAG, a preliminary run was terminated after 68 trials with 0\% FA, as the model could not complete a single compositional puzzle within the time budget.
This highlights a tension in current model design: extended thinking improves reasoning benchmarks but is counterproductive for time-constrained agentic tasks that require \emph{many shallow tool calls} rather than \emph{few deep reasoning chains}.

\section{Tool-Use Shortcuts}
\label{appendix:shortcuts}

On \aar{}-DAG, 14 to 21\% of all trials achieve the correct answer while visiting $<$30\% of required pages, compared to 6 to 11\% on \aar{}-Linear. Among correct answers specifically, shortcuts account for 45 to 58\% on \aar{}-DAG versus 16 to 30\% on \aar{}-Linear, rising to 88\% of correct answers on extreme DAG legs. If shortcuts are excluded, \aar{}-DAG accuracy drops from 31\% to 14 to 17\%, barely above the 10\% random baseline.

We do not consider this a fundamental validity threat, for three reasons. First, shortcuts are not lucky guesses: they achieve 43.8\% RCR and 60.9\% intermediate value accuracy, 3.5$\times$ above random, indicating genuine tool-chain reasoning. Second, our decomposed metrics \emph{explicitly detect} this behavior: PVR\,$<$\,0.3 flags navigation bypass. Third, shortcuts reveal a measurable property of the riddle: the clue envelope leaks enough tool-chain structure for agents to sometimes infer the answer without visiting the intended pages.

A structural analysis clarifies \emph{why} shortcuts occur. On \aar{}-DAG, 62\% of golden intermediate values belong to tool or reason stops, which can be computed through API calls and arithmetic \emph{without} visiting any Wikipedia page. The remaining 38\% are page stops that require specific Wikipedia knowledge. Shortcut agents predominantly recover tool-stop values through inferred API arguments (e.g., geocoding a location mentioned in the riddle), not recalling memorized Wikipedia facts.

Nonetheless, the high shortcut rate on extreme DAG legs (88\% of correct answers) is a limitation that inflates difficulty-level accuracy. Without shortcuts, \aar{}-DAG accuracy drops from 31\% to 14 to 17\%, underscoring benchmark difficulty when genuine navigation is required. Reducing clue leakage (e.g., more opaque phrasing) is a concrete direction for future versions, though it risks introducing ambiguity that makes puzzles unsolvable.

\section{Full Benchmark Comparison}
\label{appendix:full_comparison}

\begin{table}[H]
\centering
\setlength{\tabcolsep}{2.5pt}
\renewcommand{\arraystretch}{1.0}
\begin{tabular}{@{}lcccccccccccc@{}}
\toprule
& & & \multicolumn{4}{c}{\textit{Evaluation}} &
  \multicolumn{3}{c}{\textit{Design}} &
  \multicolumn{3}{c}{\textit{Compositionality}} \\
\cmidrule(lr){4-7}\cmidrule(lr){8-10}\cmidrule(lr){11-13}
\textbf{Benchmark} & \textbf{Venue} & \textbf{Tools}
  & \textbf{Nav} & \textbf{Met} & \textbf{Stp} & \textbf{Lve}
  & \textbf{Diff} & \textbf{Gld} & \textbf{Gen}
  & \textbf{Steps} & \textbf{\%Lin} & \textbf{\%DAG} \\
\midrule
\multicolumn{13}{l}{\textit{Tool-use \& composition}} \\
ToolBench    & ICLR'24   & 16k+  & \xmark & 2 & \xmark & \cmark$^\dagger$
             & 3 lvl & \cmark & Auto  & 1.9  & 100 & 0   \\
BFCL         & ICML'25   & 2k+   & \xmark & 3 & \xmark & \xmark
             & cat   & \cmark & Mix   & --   & --  & --  \\
TaskBench    & NeurIPS'24 & graph & \xmark & 3 & \cmark & \xmark
             & size  & \cmark & Auto  & 1.7  & 94  & 2.5 \\
T-Eval       & ACL'24    & mult  & \xmark & 6 & \cmark & \xmark
             & 2 lvl & \cmark & Man   & 4.8  & 62  & 14  \\
NESTFUL      & arXiv'24  & nest  & \xmark & 2 & \cmark & \xmark
             & depth & \cmark & Scr   & 3.4  & 55  & 45  \\
ToolHop      & ACL'25    & 3.9k  & \xmark & 1 & \xmark & \xmark
             & hops  & \cmark & Auto  & 2.9  & 100 & 0   \\
\midrule
\multicolumn{13}{l}{\textit{Web navigation \& agent}} \\
GAIA         & ICLR'24   & var   & \cmark & 1 & \xmark & \xmark
             & 3 lvl & \xmark & Man   & $\sim$5$^\ddagger$ & 100 & 0 \\
WebArena     & ICLR'24   & brow  & \cmark & 1 & \xmark & \cmark
             & impl  & \xmark & Scr   & --   & --  & --  \\
AgentBench   & ICLR'24   & 8env  & part  & 1 & \cmark & mix
             & env   & \xmark & Man   & --   & --  & --  \\
AgentBoard   & NeurIPS'24 & 9env & part  & 2 & \cmark & mix
             & sub   & \xmark & Man   & --   & --  & --  \\
AppWorld     & ACL'24    & 457   & \xmark & 1 & \xmark & \xmark
             & 2 lvl & \xmark & Man   & --   & --  & --  \\
tau-bench    & arXiv'24  & dom   & \xmark & 1 & \cmark & \xmark
             & 2 dom & \cmark & Man   & --   & --  & --  \\
\midrule
\textbf{\aar{}} & --     & \textbf{19} & \cmark & \textbf{3} & \cmark & \cmark
             & \textbf{4 lvl} & \cmark & \textbf{Auto} & \textbf{22.1} & \textbf{0} & \textbf{100} \\
\bottomrule
\end{tabular}
\caption{Full comparison with 12 representative benchmarks (condensed version in Table~\ref{tab:comparison}).
\textbf{Nav}: navigation required.
\textbf{Met}: number of metrics.
\textbf{Stp}: step-level evaluation.
\textbf{Lve}: live API data ($^\dagger$ToolBench suffers instability).
\textbf{Diff}: difficulty control.
\textbf{Gld}: verified gold trace.
\textbf{Gen}: generation method (Auto/Man/Scr/Mix).
\textbf{Steps}: mean pit stops per golden chain.
\textbf{\%Lin}/\textbf{\%DAG}: fraction of strictly linear vs.\ branching instances.
$^\ddagger$GAIA step count from annotator metadata (validation split only).}
\label{tab:comparison_full}
\end{table}

\section{Full Results by Benchmark Variant}
\label{appendix:per_difficulty_figs}

\begin{figure}[H]
\centering
\includegraphics[width=\linewidth]{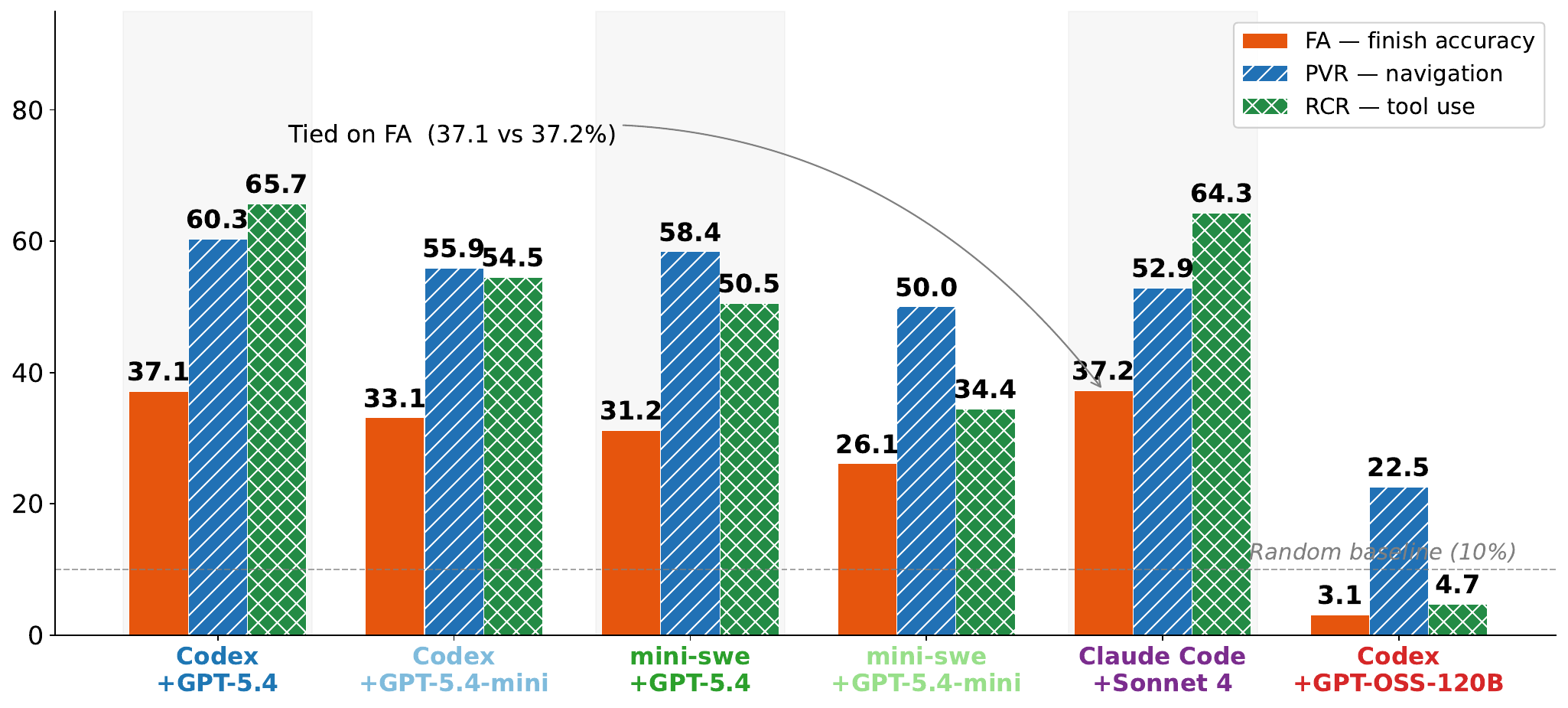}
\caption{Main results on both benchmark variants: (a)~\aar{}-Linear (800 legs, 6 configs including GPT-OSS-120B), (b)~\aar{}-DAG (600 legs, 5 configs). PVR drops 13 to 18pp from Linear to DAG while RCR remains stable or increases.}
\label{fig:overview_full}
\end{figure}

\begin{figure}[H]
  \centering
  \begin{subfigure}[t]{0.48\linewidth}
    \includegraphics[width=\linewidth]{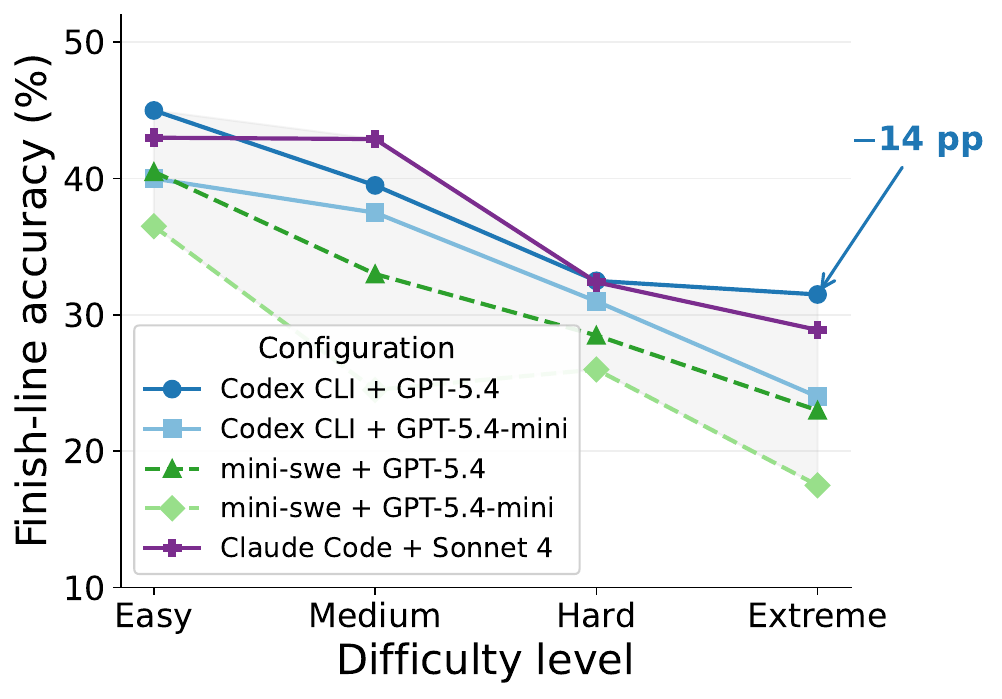}
    \caption{FA degrades monotonically (best: $-$13.5\,pp, worst: $-$19.0\,pp).}
    \label{fig:fa-difficulty}
  \end{subfigure}
  \hfill
  \begin{subfigure}[t]{0.48\linewidth}
    \includegraphics[width=\linewidth]{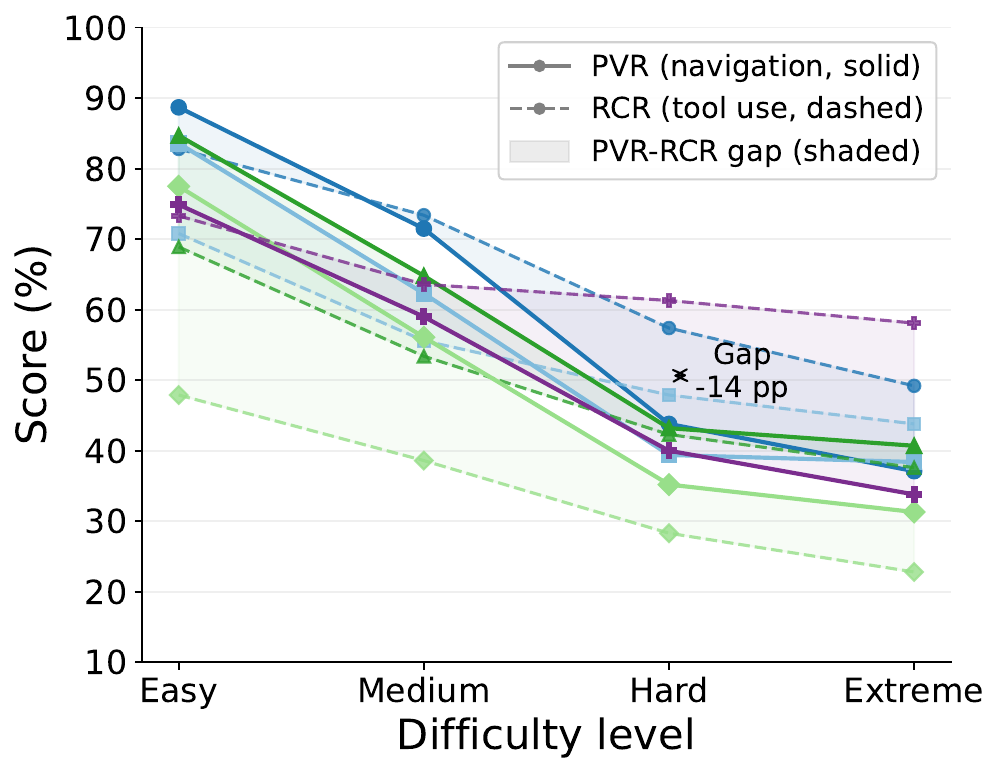}
    \caption{PVR (solid) falls $2\times$ faster than RCR (dashed).}
    \label{fig:pvr-rcr-gap}
  \end{subfigure}
  \caption{Per-difficulty breakdown on \aar{}-Linear. Navigation quality degrades far faster than tool-use competence.}
  \label{fig:findings12}
\end{figure}

\section{Full Results Table}
\label{appendix:full_results}

\begin{table*}[ht]
\centering
\small
\setlength{\tabcolsep}{3.5pt}
\begin{tabular}{@{}lll ccc ccc@{}}
\toprule
& & & \multicolumn{3}{c}{\textbf{\aar{}-Linear} (800 legs)} & \multicolumn{3}{c}{\textbf{\aar{}-DAG} (600 legs)} \\
\cmidrule(lr){4-6} \cmidrule(lr){7-9}
\textbf{Agent} & \textbf{Model} & \textbf{Level} & \textbf{FA} & \textbf{PVR} & \textbf{RCR} & \textbf{FA} & \textbf{PVR} & \textbf{RCR} \\
\midrule
\multirow{5}{*}{Codex CLI}
& \multirow{5}{*}{GPT-5.4}
& Easy    & 45.0 & 88.7 & 82.8 & 26.0 & 76.1 & 86.8 \\
& & Medium  & 39.5 & 71.5 & 73.4 & 30.0 & 55.5 & 75.5 \\
& & Hard    & 32.5 & 43.8 & 57.4 & 31.9 & 32.6 & 64.0 \\
& & Extreme & 31.5 & 37.1 & 49.2 & 35.9 & 24.3 & 55.3 \\
& & \cellcolor{gray!10}\textit{All} & \cellcolor{gray!10}\textit{37.1} & \cellcolor{gray!10}\textit{60.3} & \cellcolor{gray!10}\textit{65.7} & \cellcolor{gray!10}\textit{31.7} & \cellcolor{gray!10}\textit{43.0} & \cellcolor{gray!10}\textit{68.0} \\
\midrule
\multirow{5}{*}{Codex CLI}
& \multirow{5}{*}{GPT-5.4-mini}
& Easy    & 40.0 & 83.6 & 70.8 & 26.0 & 62.9 & 71.2 \\
& & Medium  & 37.5 & 62.3 & 55.6 & 30.0 & 47.7 & 58.5 \\
& & Hard    & 31.0 & 39.4 & 47.9 & 29.5 & 29.2 & 54.9 \\
& & Extreme & 24.0 & 38.4 & 43.8 & 34.8 & 22.8 & 47.6 \\
& & \cellcolor{gray!10}\textit{All} & \cellcolor{gray!10}\textit{33.1} & \cellcolor{gray!10}\textit{55.9} & \cellcolor{gray!10}\textit{54.5} & \cellcolor{gray!10}\textit{30.7} & \cellcolor{gray!10}\textit{37.5} & \cellcolor{gray!10}\textit{56.3} \\
\midrule
\multirow{5}{*}{mini-swe-agent}
& \multirow{5}{*}{GPT-5.4}
& Easy    & 40.5 & 84.7 & 68.9 & 26.0 & 69.8 & 66.4 \\
& & Medium  & 33.0 & 64.8 & 53.4 & 30.0 & 52.0 & 56.2 \\
& & Hard    & 28.5 & 43.2 & 42.3 & 25.9 & 31.7 & 42.5 \\
& & Extreme & 23.0 & 40.7 & 37.6 & 34.2 & 28.4 & 37.8 \\
& & \cellcolor{gray!10}\textit{All} & \cellcolor{gray!10}\textit{31.2} & \cellcolor{gray!10}\textit{58.4} & \cellcolor{gray!10}\textit{50.5} & \cellcolor{gray!10}\textit{29.5} & \cellcolor{gray!10}\textit{42.1} & \cellcolor{gray!10}\textit{48.5} \\
\midrule
\multirow{5}{*}{mini-swe-agent}
& \multirow{5}{*}{GPT-5.4-mini}
& Easy    & 36.5 & 77.5 & 47.9 & 24.0 & 63.6 & 52.3 \\
& & Medium  & 24.5 & 56.1 & 38.6 & 28.0 & 42.6 & 40.4 \\
& & Hard    & 26.0 & 35.2 & 28.3 & 25.9 & 26.7 & 27.0 \\
& & Extreme & 17.5 & 31.3 & 22.8 & 34.2 & 21.7 & 20.7 \\
& & \cellcolor{gray!10}\textit{All} & \cellcolor{gray!10}\textit{26.1} & \cellcolor{gray!10}\textit{50.0} & \cellcolor{gray!10}\textit{34.4} & \cellcolor{gray!10}\textit{28.7} & \cellcolor{gray!10}\textit{35.3} & \cellcolor{gray!10}\textit{32.6} \\
\midrule
\multirow{5}{*}{Claude Code}
& \multirow{5}{*}{Sonnet 4}
& Easy    & 43.0 & 74.9 & 73.3 & 27.5 & 68.7 & 77.5 \\
& & Medium  & 42.9 & 59.0 & 63.6 & 33.9 & 45.3 & 79.3 \\
& & Hard    & 32.4 & 40.0 & 61.3 & 36.4 & 25.1 & 70.2 \\
& & Extreme & 28.9 & 33.8 & 58.1 & 42.0 & 26.4 & 62.3 \\
& & \cellcolor{gray!10}\textit{All} & \cellcolor{gray!10}\textit{37.2} & \cellcolor{gray!10}\textit{52.9} & \cellcolor{gray!10}\textit{64.3} & \cellcolor{gray!10}\textit{35.8} & \cellcolor{gray!10}\textit{38.7} & \cellcolor{gray!10}\textit{71.6} \\
\midrule
\multicolumn{3}{l}{\textit{Non-agent baselines}} \\
Random & -- & All & 10.0 & 0.0 & 0.0 & 10.0 & 0.0 & 0.0 \\
\bottomrule
\end{tabular}
\caption{Main results (\%) on both benchmark variants. \textbf{FA}: finish-line accuracy. \textbf{PVR}: pit-stop visit rate. \textbf{RCR}: roadblock completion rate.}
\label{tab:main_results_appenidx}
\end{table*}

\section{Computational Resources}
\label{appendix:resources}

Table~\ref{tab:resources} summarizes the computational resources for the full evaluation. Token usage varies by an order of magnitude across agent frameworks: Codex CLI averages 1.4--1.8M tokens/trial due to its extensive planning loops, while mini-swe-agent uses only 149K--187K tokens/trial.
Claude Code uses fewer tokens than both (114--225K/trial), yet takes the longest wall-clock time (292--320s), reflecting a deliberate approach with targeted tool calls and iterative error recovery.
Despite achieving comparable accuracy (37.2\% vs.\ 37.1\% on \aar{}-Linear), Claude Code consumes 6$\times$ fewer tokens than Codex CLI, suggesting that token efficiency and task performance are largely decoupled.
Across all 7{,}000 trials (10 configurations $\times$ 600--800 legs), the evaluation consumed 286 compute-hours.

\begin{table}[H]
\centering
\setlength{\tabcolsep}{2pt}
\begin{tabular}{@{}ll r r r@{}}
\toprule
\textbf{Agent} & \textbf{Model} & \textbf{Tok.} & \textbf{Time} & \textbf{Total} \\
\midrule
\multicolumn{5}{l}{\textit{\aar{}-Linear (800 legs per config)}} \\
Codex CLI    & GPT-5.4      & 1.44M & 211\,s & 46.9\,h \\
Codex CLI    & GPT-5.4-mini & 1.66M &  92\,s & 20.4\,h \\
mini-swe     & GPT-5.4      & 154K  &  58\,s & 12.8\,h \\
mini-swe     & GPT-5.4-mini & 156K  &  31\,s &  6.9\,h \\
Claude Code  & Sonnet 4     & 225K  & 292\,s & 65.0\,h \\
\midrule
\multicolumn{5}{l}{\textit{\aar{}-DAG (600 legs per config)}} \\
Codex CLI    & GPT-5.4      & 1.79M & 260\,s & 43.3\,h \\
Codex CLI    & GPT-5.4-mini & 1.76M & 119\,s & 19.8\,h \\
mini-swe     & GPT-5.4      & 187K  &  71\,s & 11.8\,h \\
mini-swe     & GPT-5.4-mini & 149K  &  33\,s &  5.5\,h \\
Claude Code  & Sonnet 4     & 114K  & 320\,s & 53.4\,h \\
\midrule
\multicolumn{3}{l}{\textit{Grand total (7{,}000 trials)}} & & 286.0\,h \\
\bottomrule
\end{tabular}
\caption{Computational resources per configuration. \textbf{Tok.}: mean input$+$output tokens per trial. \textbf{Time}: mean wall-clock time per trial. \textbf{Total}: cumulative agent time.}
\label{tab:resources}
\end{table}

\end{document}